\documentclass[10pt,twocolumn,letterpaper]{article}

\usepackage[pagenumbers]{cvpr} %

\usepackage{graphicx}
\usepackage{amsmath}
\usepackage{amssymb}
\usepackage{booktabs}
\usepackage{multirow}
\usepackage{amsthm}
\usepackage{enumitem}
\usepackage{needspace}

\usepackage{csquotes}

\usepackage[export]{adjustbox}
\usepackage{pgfplots}
\usepackage[most]{tcolorbox}
\usepackage{caption}
\usepackage{subcaption}
\usepackage{array}
\usepackage{etoolbox}
\usepackage{xfrac}
\usepackage[normalem]{ulem}
\usepackage{placeins}
\usepackage{pifont}
\usepackage{tabularx}
\usepackage{microtype}
\usepackage[outline]{contour}
\usepackage[absolute,overlay]{textpos}
\usepackage{tikz}
\usepackage{tikzit}
\usepackage{mwe}

\usetikzlibrary{positioning,calc,arrows.meta}
\usetikzlibrary{shapes.geometric, shadings, patterns}
\usepgfplotslibrary{colorbrewer}
\pgfplotsset{compat=1.18}

\usepackage{tablefootnote}
\usepackage{pgf}

\newtheoremstyle{remarkstyle} %
  {3pt} %
  {3pt} %
  {} %
  {} %
  {\bfseries} %
  {.} %
  { } %
  {\thmname{#1}} %
  
\theoremstyle{remarkstyle}

\usepackage{amsmath}
\usepackage{booktabs}
\usepackage{algorithm}
\usepackage{algpseudocode}

\definecolor{cvprblue}{rgb}{0.21,0.49,0.74}

\usepackage[pagebackref,breaklinks,colorlinks,allcolors=cvprblue]{hyperref}
\usepackage[capitalize]{cleveref}
\crefname{section}{Sec.}{Secs.}
\Crefname{section}{Section}{Sections}
\Crefname{table}{Table}{Tables}
\crefname{table}{Tab.}{Tabs.}

\usepackage{paralist, tabularx}

\usepackage{pifont}%
\newcommand{\cgreenmark}{{\color{green} \ding{52}}}%
\newcommand{\xredmark}{{\color{red} \ding{55}}}%

\usepackage{twemojis}
\usepackage{fontawesome5}

\definecolor{baselinecolor}{RGB}{0, 102, 204} %
\definecolor{synthecolor}{RGB}{128, 128, 128} %
\definecolor{flaircolor}{RGB}{204, 0, 0} %
\definecolor{LEVIRCOLOR}{RGB}{30, 150, 252}
\definecolor{SECONDCOLOR}{RGB}{184, 146, 255}
\definecolor{S2COLOR}{RGB}{144, 190, 109}
\definecolor{HIUCDCOLOR}{RGB}{249, 65, 68}
\definecolor{ANYCOLOR}{RGB}{75,0,130}
\definecolor{blueaccent}{RGB}{0,150,214}
\definecolor{greenaccent}{RGB}{0,139,43}
\definecolor{purpleaccent}{RGB}{130,41,128}
\definecolor{orangeaccent}{RGB}{240,83,50}

\definecolor{baselinecolor}{RGB}{0,150,214}
\definecolor{CPcolor}{RGB}{0,139,43}
\definecolor{SyntheColor}{RGB}{130,41,128}
\definecolor{FCcolor}{RGB}{240,83,50}

\def\paperID{1372}
\def\confName{CVPR}
\def\confYear{2025}

\newcommand{\PIPELINE}{\texttt{HySCDG}}
\newcommand{\FLAIRCHANGE}{FSC-180k} %
\newcommand{\FLAIRChange}{FSC-180k} %
\newcommand{\imgwidth}{2.8cm}
\newcommand{\imgwidthl}{3cm}
\newcommand{\imgspace}{0.05cm}
\newcommand{\rowspace}{0.1cm}

\newcommand{\clem}[1]{\textcolor{brown}{ #1}}

\begin{document}

\title{The Change You Want To Detect:\\ Semantic Change Detection In Earth Observation With Hybrid Data Generation}

\author{Yanis Benidir
\and 
Nicolas Gonthier\\
Univ Gustave Eiffel, ENSG, IGN, LASTIG, France \\
{\tt\small firstname.lastname@ign.fr}
\and 
Clément Mallet
}

\maketitle

\begin{abstract}
Bi-temporal change detection at scale based on Very High Resolution (VHR) images is crucial for Earth monitoring. This remains poorly addressed so far: methods either require large volumes of annotated data (semantic case), or are limited to restricted datasets (binary set-ups). Most  approaches do not exhibit the versatility required for temporal and spatial adaptation: simplicity in architecture design and pretraining on realistic and comprehensive datasets. Synthetic datasets are the key solution but still fail to handle complex and diverse scenes.
In this paper, we present \PIPELINE~ a generative pipeline for creating a large hybrid semantic change detection dataset that contains both real VHR images and inpainted ones, along with land cover semantic map at both dates and the change map.
Being semantically and spatially guided, \PIPELINE~ generates realistic images, leading to a comprehensive and hybrid transfer-proof dataset \FLAIRCHANGE.
We evaluate \FLAIRCHANGE~on five change detection cases (binary and semantic), from zero-shot to mixed and sequential training, and also under low data regime training. Experiments demonstrate that pretraining on our hybrid dataset leads to a significant performance boost, outperforming SyntheWorld, a fully synthetic dataset, in every configuration. All codes, models, and data are available here: \href{https://yb23.github.io/projects/cywd/}{https://yb23.github.io/projects/cywd/}.

\end{abstract}

\section{Introduction}
\begin{figure}[ht!]
    \centering
   \includegraphics[width=0.9\linewidth]{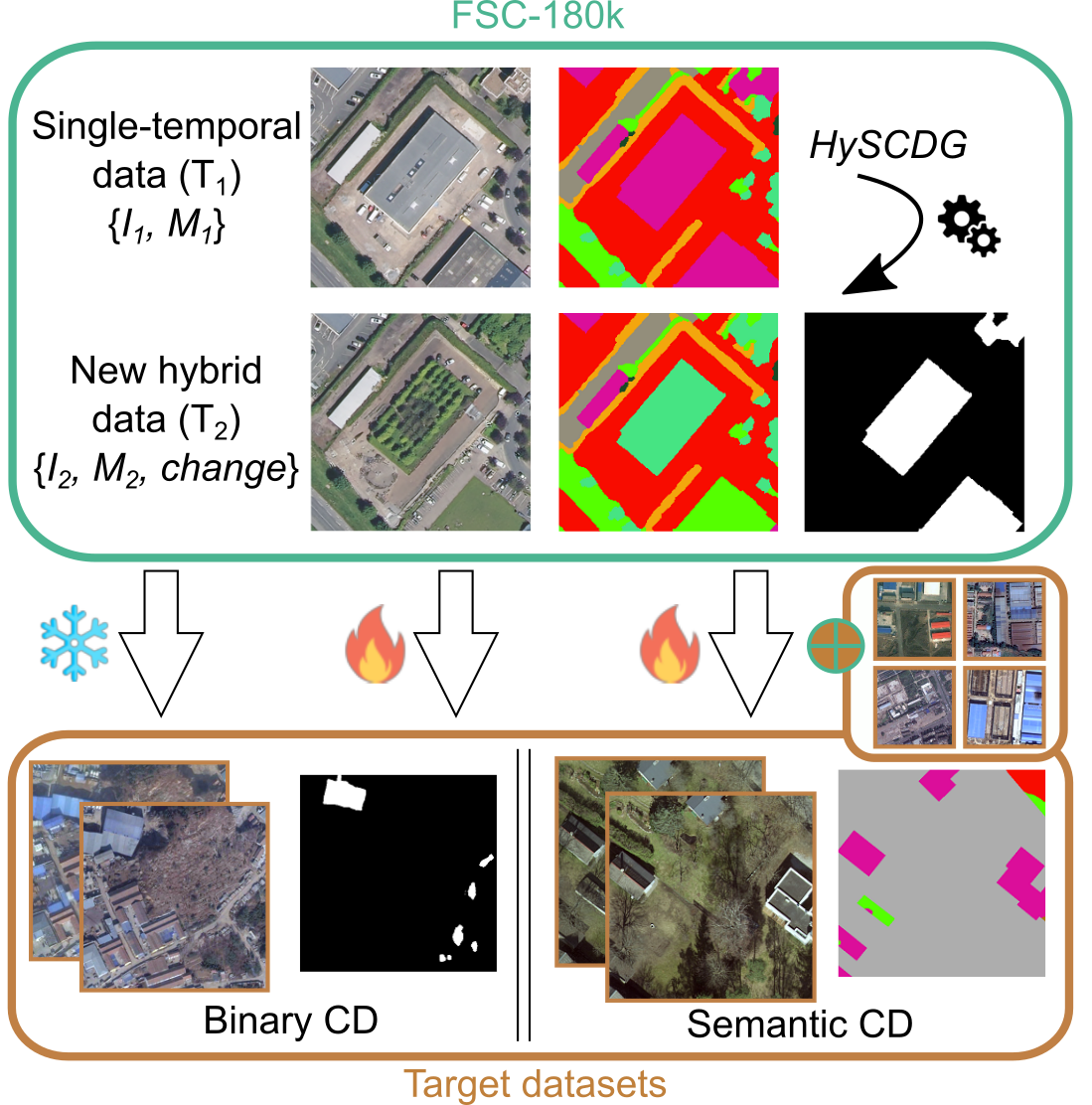}
    \caption{{\bf Efficient and scalable change detection} requires a comprehensive training dataset that does not exist today. Using a single-temporal dataset (image+semantic map), we propose \PIPELINE~ that generates \textbf{a novel bi-temporal hybrid dataset} \FLAIRChange. This enables multiple transfer learning possibilities on either binary and semantic change detection tasks. \includegraphics[width=0.04\linewidth]{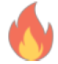}/ \includegraphics[width=0.04\linewidth]{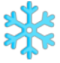}: Transfer with/without fine-tuning our model, respectively.}
    \label{fig:teaser}
\end{figure}

Efficient methods for change detection (CD) are crucial for monitoring territories and the various phenomena and activities that impact them \cite{Burke_Science}. As human and climatic pressures increase, this topic is of utmost importance and calls every year for higher spatial resolution and semantic accuracy \cite{Nature_swin}.
Change detection at scale has benefited greatly from the contributions of deep learning in recent years \cite{daudt2018fully,ZHENG2024239,zheng2024changen2,ChangeMamba}, but most methods are grounded on large volumes of annotated data \cite{Van_Etten_2021_CVPR, Priya_fire}. 
Creating a large-scale dataset for bi-temporal (pair-wise) remote sensing change detection poses significant challenges and costs, especially when dealing with Very High Resolution (VHR) images (ground sampling distance: GSD $<$1$\:$m, \cite{Li_2024_CVPR}). This is due to the need for expertise and considerable effort in collecting, preprocessing, and annotating images \cite{Toker_2022_CVPR,xia2023openearthmap}, particularly in the case of VHR data and multiple labels (termed \textit{semantic change detection}, SCD) \cite{Zheng_2021_ICCV, ZHU2022113266,TIAN2022164,WANG2024103761}.

As an alternative, synthetic data generation is a promising direction for bi-temporal labeled data provision from a single temporal domain that is successful in diverse areas such as urban sprawl \cite{SMARS}, disaster or climate change monitoring \cite{9420150} or crop mapping \cite{ijgi12110450}. Two main paradigms exist for CD: fully synthetic datasets (mainly graphics-based \cite{kolos2019procedural,SyntheWorld}), and hybrid solutions that mix real and modified images, often at the instance level \cite{chen2021adversarial,Changen,zheng2024changen2}, known as \textit{data augmentation}. The first method synthesizes image pairs by rendering two versions of a 3D virtual world thanks to a computer graphics engine.
On the other hand, hybrid approaches synthesize new image pairs by pasting or erasing object instances into existing images (\textit{inpainting}). They propose a suitable trade-off and faster generation of large quantities of data, with likely higher realism. However, none of these  solutions meet the requirements for scalable SCD with VHR imagery (\cref{tab:synthetic_datasets}): focus on single tasks, restricted geographical areas, limited diversity, no guarantee on semantic consistency between pairs \cite{SEO_wacv} and subsequently poor transfer to real datasets. 
To address this challenge, we introduce a generative pipeline built upon Stable Diffusion \cite{LatentDiffusionModels} and ControlNet \cite{ControlNet}, designed to harness an existing VHR semantic land cover dataset \cite{garioud2023flair} and instance masks, to generate an extensive hybrid semantic bi-temporal change detection dataset encompassing both real and inpainted images, semantic land cover maps at each date, and a change map (\cref{fig:teaser}). Then, we evaluate the transferability of our synthetic dataset on five datasets for both binary and semantic change detection cases ranging from zero-shot to sequential and mixed training set-ups, as well as under low-data regime.  

Our contributions are as follows:
\begin{compactitem} %
    \item We present a \textbf{generation pipeline} \emph{\PIPELINE}~ for creating VHR bi-temporal SCD datasets based on a land cover real-world dataset by using Stable Diffusion and ControlNet with tunable and semantically-guided inpainting of individual objects. It can be easily configured or finetuned on any land-cover dataset,  generating images with the dataset’s distinctive traits.
    \item We provide \emph{\FLAIRChange}: a \textbf{Hybrid Semantic Change Detection pretraining dataset}, created from FLAIR \cite{garioud2023flair} along with the instance masks for 300k objects for the corresponding images.
    \item We provide a comprehensive evaluation of the proposed synthetic dataset on multiple transfer learning scenarios on five change datasets, in particular for SCD.
\end{compactitem}

\section{Related work}
\label{sec:related}

\paragraph{Detecting changes in the deep learning era.}

We target bi-temporal land cover change detection in VHR images \cite{lv2021land}. 
This historical task \cite{howarth1981procedures,singh1989review,nemmour2006multiple} consists in detecting the \textit{meaningful} elements of a given area that have changed between two acquisitions and providing the land cover categories at each period (also known as \textit{change trajectory}). It presents challenges: radiometric variations, geometric variations and non permanent changes. They have been extensively addressed with supervised deep learning techniques \cite{Zheng_2021_ICCV}: from CNN \cite{daudt2018fully,zhan2017change,daudt2018hrscd} to Transformers \cite{ZHENG2022228,ZHENG2024239,ChangeMamba,ChangeFormer,SCanNet,Nature_swin}, mainly with Siamese architecture. 
Multiple datasets, accompanied with ad-hoc methods, have been created to feed such models. They are often limited by their size \cite{hiucd,LevirCD,SECOND,pang2023detecting,ji2018fully_WHU}, geographical extent \cite{LevirCD,SECOND,hiucd,TIAN2022164}, low geometrical or annotation quality \cite{daudt2018hrscd}. Also, most of the effort focuses on binary change detection, \textit{i.e.}, restricted to a unique category, mainly \textit{buildings} \cite{LevirCD,s2looking,pang2023detecting,ji2018fully_WHU,peng2020semicdnet} or \textit{disasters} \cite{ZHANG20231}.
Such scarcity in high quality, diverse, large-scale labels is alleviated with transfer learning \cite{shi2020change,CAO2023113371}, by fine-tuning models pretrained on large-scale datasets for different image domains, even for semantic change detection  \cite{SECOND,hiucd,TIAN2022164}.

\paragraph{Synthesizing and inpainting remote sensing images.}
\textit{Synthetic} remote sensing data is useful \textit{e.g.,} for cloud removal \cite{czerkawski2022deep}, image restoration \cite{fan2011pixel}, or training deep supervised models \cite{toker2024satsynth}, before transfer to real examples. %
Solutions based on patches, \cite{meng2017sparse,zheng2021nonlocal}, autoencoders \cite{huang2022image}, GAN  \cite{marin2021sssgan,kuznetsov2020remote}, pixel-aligned generation \cite{isola2017image} have been proposed, but with limited image quality and no semantic control.
Diffusion models (DM) have improved such quality \cite{LatentDiffusionModels}, but most models have applied DM directly on mid-resolution RGB images \cite{khanna2023diffusionsat,GenerateYourOwnScotland,toker2024satsynth}, neglecting both the multi-spectral dimension  \cite{czerkawski2024exploring} and VHR spaces.  
However, a key finding remains: semantically control the synthesis, \textit{e.g.}, by directly synthesizing text \cite{khanna2023diffusionsat}, semantic maps \cite{toker2024satsynth} or with a control module guided by edges \cite{czerkawski2024exploring}, semantic maps \cite{GenerateYourOwnScotland} or metadata \cite{khanna2023diffusionsat}. 
In our work, we fine-tune a Stable Diffusion and a ControlNet models to achieve semantically-controlled inpainting of VHR images as a basis to take advantage of existing images, and generate an hybrid semantic change detection dataset at scale and with high diversity.

\paragraph{Generating synthetic change datasets.}

Two main approaches exist: fully synthetic or hybrid solutions. 
The first paradigm leverages advanced 3D rendering engines to create fully synthetic datasets \cite{kolos2019procedural,SMARS}. This procedural approach allows complete control over rendering parameters (instance placement, semantic classes, lighting conditions, etc.) generated from scratch.
\textit{E.g.}, SyntheWorld \cite{SyntheWorld} effectively leverages DM prompted by GPT-4 to create a diverse and large-scale fully synthetic dataset. 
On the other hand, hybrid approaches generate (semi-)synthetic data by inserting fake changes into real images. 

Such modifications can be random crops \cite{SelfPair}, instance copy-pasting, or inpainting made by GAN \cite{SelfPair,Changen,chen2021adversarial} or DM \cite{zheng2024changen2,tang2024changeanywhere}. 
 Blending or style transfer methods are often used to enhance diversity \cite{chen2021adversarial,SelfPair}.\\
 A pivotal challenge is the semantic guidance: modifying few elements of images to mimic real scarce changes, \textit{e.g.}, with instance footprints extracted from external datasets \cite{chen2021adversarial} or segmentation model outputs \cite{zheng2024changen2}. %
\cite{tang2024changeanywhere} adopt a Denoising diffusion probabilistic model to generate a new image directly from a modified semantic map and a degraded version of the corresponding snapshot.
Changen2 \cite{zheng2024changen2} is the closest work to ours, with a full pipeline to synthesize a hybrid semantic dataset, again with a DM conditioned by a semantic map. However, Changen2 creates synthetic data specific to a given target dataset in terms of image resolution, size, or change characteristics.
In this paper, we explore semantically controlled inpainting based on DM and ControlNet to the intertwined challenge of diverse and multi-class modification of VHR images. It is termed \PIPELINE~ and leads to the hybrid dataset named \FLAIRCHANGE.  \cref{tab:synthetic_datasets} provides a comparison with main synthetic change detection datasets.%

\begin{table}[ht!]
\caption{\textbf{Synthetic Change Datasets.} Land cover change detection in remote sensing imagery. For the total number of pixels, we only consider strictly different images. The geographical zone's extension can be local \faBuilding[regular], national \twemoji{flag: France} or global \faGlobe. We denote
open-access (OA) and directly downloadable datasets with \cgreenmark.
In the case of IAug, we sum the number of pixels of the two datasets provided in this work. \textit{OEM} means \textit{OpenEarthMap}.}
\label{tab:synthetic_datasets} %
\centering
\scriptsize
\renewcommand{\arraystretch}{1.5}
\setlength{\tabcolsep}{2pt}
\resizebox{\columnwidth}{!}{ 
\begin{tabular}{lrrrclll}
\toprule
Dataset & OA &  Pixels $\times 10^6$ & GSD (m) & Classes  & Source & Zone &  Type \\ \midrule
SynCW \cite{kolos2019procedural} & \xredmark  & 37 & 0.6& 1 & X &   \faBuilding[regular]  & Synthetic \\
SMARS \cite{SMARS} & \cgreenmark & 110 & 0.3-0.5& 2 &  X & \faBuilding[regular]   & Synthetic \\%link https://www.dlr.de/en/eoc/about-us/remote-sensing-technology-institute/photogrammetry-and-image-analysis/public-datasets/smars 
IAug \cite{chen2021adversarial} & \xredmark  &  1,167 & 0.075-0.5 & 1 & \begin{tabular}{@{}c@{}}LEVIR-CD \cite{LevirCD}, \\ WHU-CD \cite{ji2018fully_WHU}\end{tabular} & \faBuilding[regular]   & Hybrid \\
Ce-100K \cite{tang2024changeanywhere} & \xredmark  & 6,553 & 0.25-0.5& 8 & OEM \cite{xia2023openearthmap} & \faGlobe & Synthetic  \\
Changen2 \cite{zheng2024changen2} & \xredmark & 7,077 & 0.25–0.5& 8 & OEM \cite{xia2023openearthmap} & \faGlobe & Hybrid \\ %
Changen \cite{Changen} & \xredmark & 11,796 & $~$ 0.8& 1 & xView2 \cite{ritwik2019xbd}  & \faBuilding[regular]    & Hybrid \\
SyntheWorld \cite{SyntheWorld} & \cgreenmark & 18,350 & 0.3-1& 1 & X & X & Synthetic \\
\midrule
\textbf{\FLAIRCHANGE~(Ours)} &  \cgreenmark & \textbf{80,740} & \textbf{0.2} & \textbf{16} & FLAIR \cite{garioud2023flair} & \twemoji{flag: France}  &  Hybrid \\ 
\bottomrule
\end{tabular}
}
\end{table}

\paragraph{Transfer learning from synthetic/hybrid datasets.}
Transfer learning is used to assess the validity of the generated image pairs, and vary with the operational set-up \cite{ma2024transfer}. SCD at scale is performed through sequential learning: pretraining on such datasets and fine-tuning on real-world cases \cite{Changen,zheng2024changen2,kolos2019procedural,chen2021adversarial}. 
Fewer works \cite{SyntheWorld} evaluate on mixed training sets (single training mixing both real and synthetic samples), which can help to prevent overfitting. 
An orthogonal solution considers low data regimes, \textit{i.e.}, fine-tuning on a very small train set \cite{SyntheWorld,tang2024changeanywhere,chen2021adversarial}, or even under zero-shot scenarios \cite{tang2024changeanywhere}.  
We evaluate our proposed dataset in the four conceivable transfer scenarios.

\section{Hybrid generation of semantic changes}
\label{sec:method}
Much of the success of deep learning methods for Earth observation relies on large-scale training datasets with reliable annotations \cite{garioud2023flair,roscher2023better}. However, no large-scale publicly available dataset exists for bi-temporal SCD: we propose \PIPELINE~for \textbf{Hybrid Semantic Change Detection Generation}, a procedure to leverage existing large-scale land cover datasets and state-of-the-art image inpainting methods to simulate visually realistic bi-temporal semantic changes. This leads to the generation of the \FLAIRChange~dataset.

\begin{figure*}[t]
    \centering
     \includegraphics[width=0.85\linewidth]{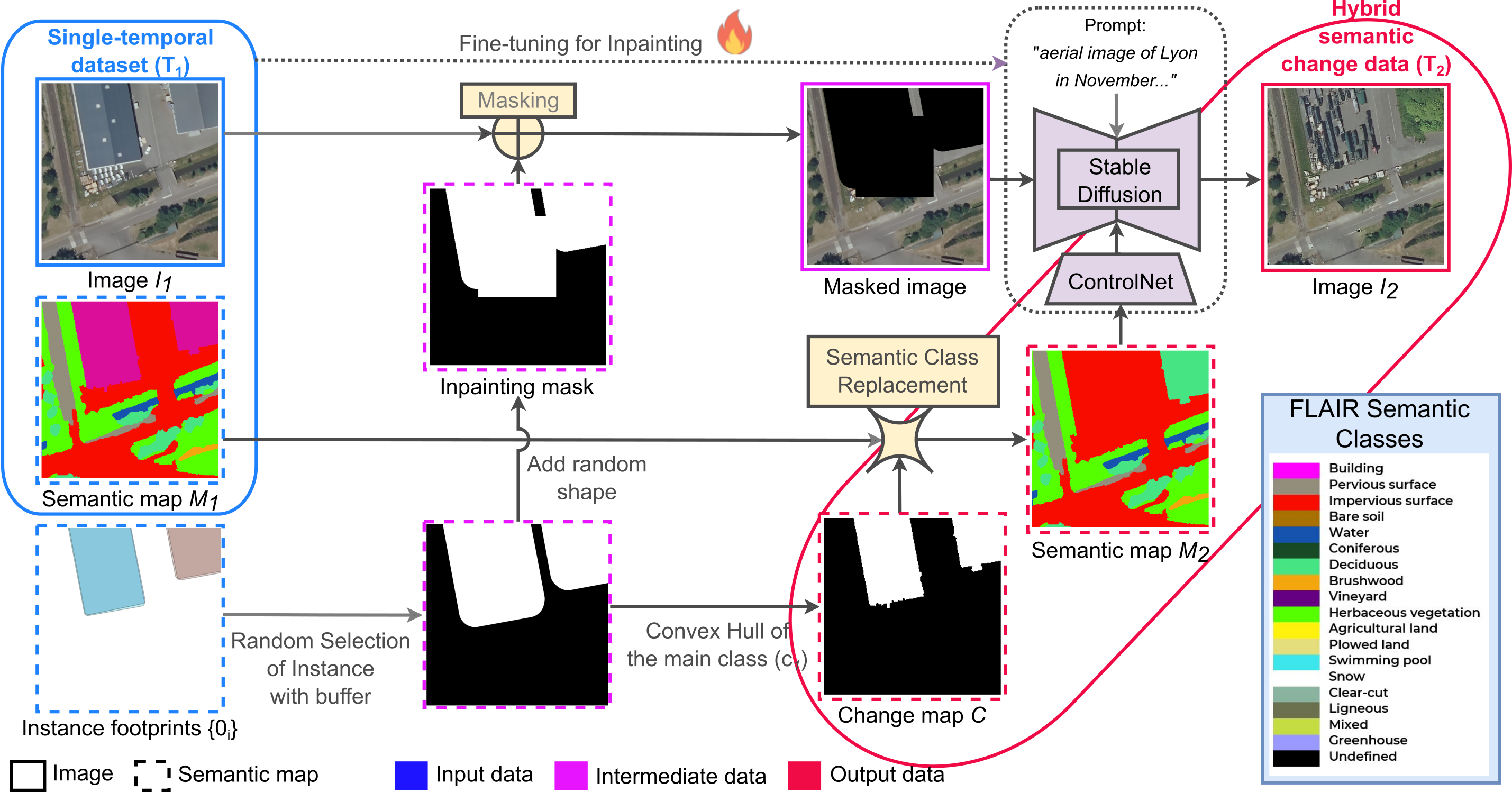}
    \caption{{\bf \PIPELINE~pipeline.} From a single-temporal dataset composed of one VHR image $I_1$, a semantic map $M_1$, and some openly available labeled instances, we generate a new VHR image $I_2$, a new map $M_2$ and subsequently a change map $C$. This results in the \FLAIRCHANGE~ hybrid dataset. The two pivotal novelties consists in: (i) \textbf{Adapting and fine-tuning a Stable-Diffusion  Model from image inpainting }and (ii) \textbf{exploiting open geospatial data }for inpainting prompt control and semantically guiding the objects to be modified. The combination of both solutions ensures a\textbf{ diverse at-scale VHR multi-class change detection} dataset. }
    \label{fig:generation_pipeline}
\end{figure*}

Our paper is grounded on two main ideas: (i) \textbf{Adapting and fine-tuning a Stable Diffusion Model} enables efficient inpainting of VHR images with a semantic control and geographical instance selection; (ii) Changes can be simulated with \textbf{sufficient diversity by randomly selecting instances (and modifying their labels) within an existing land cover map} that is paired with the image used as input.
The pipeline is shown in \cref{fig:generation_pipeline}. Details are provided in the following.

\subsection{Adaptation of Stable Diffusion for Inpainting.}

We use Stable Diffusion (SD) \cite{LatentDiffusionModels}, an open and state-of-the-art text-to-image synthesis method. 
Our objective is to modify SD to manage controllable inpainting specifically in remote sensing aerial images, for various landscapes or semantic classes. Though SD models are effective in generating natural images with clear focal points and layered depth, with artistic or cinematic styles, they require adaptation to be used for remote sensing images \cite{khanna2023diffusionsat}: acquisition distance, intricate textures, no domination of a single element. 
Starting from a Stable Diffusion 2 Inpainting checkpoint\footnote{ \href{https://huggingface.co/stabilityai/stable-diffusion-2-inpainting}{https://huggingface.co/stabilityai/stable-diffusion-2-inpainting}}, we sequentially fine-tune the Autoencoder and the Diffusion U-Net, before adding and training a ControlNet \cite{ControlNet} that leverages information extracted from an existing large-scale land cover dataset (namely, FLAIR \cite{garioud2023flair}, see \cref{subsec:pipeline}).
Our pipeline allows to generate \textbf{5-band images} that incorporate infrared and elevation along with RGB channels, for the first time in satellite image synthesis.

\paragraph{Adaptation of the Variational Autoencoder.}
Unlike most SD extensions, which focus on fine-tuning the central U-Net, we also adapted the Variational Autoencoder (VAE) that encodes images into the diffusion latent space. This adaptation targets efficient encoding of multi-band VHR images from the FLAIR dataset. The original VAE, trained with a KL-regularization factor \cite{TamingTansformers}, uses a combination of L1 and perceptual loss \cite{PerceptualLoss} alongside KL-divergence. To improve compression of remote sensing images, we added both L2 and Focal losses \cite{focal}: minimizing high-frequency errors reduces blurriness. Additionally, we included a \enquote{color loss} (L2 loss on each 5$\times$5 patch) to limit oversaturation, which affects ground object color realism. Fine-tuning of the VAE took 160 hours A100 GPU.

\paragraph{Stable Diffusion training.}
We then fine-tuned the Diffusion U-Net for inpainting, adapting it to the new VAE and teaching it to synthesize and inpaint VHR images. Special attention was given to prompt selection. Although inpainting is the model's primary function, we included direct image generation from prompts in 10\% to 20\% of cases, as in standard SD. This training required 300 hours on  a A100 GPU (30,000 iterations, batch size of 32).

\paragraph{Prompt Engineering.}
Prompts are designed using both each image's geographic coordinates and open land cover data (\textit{OpenStreetMap} \cite{OpenStreetMap}). To avoid learning irrelevant details, prompts are composed of three elements: (i)  Spatial (place name, city, region), (ii) Temporal (time of day, season), (iii) Semantic (frequent classes in the inpainted area, compared to the overall dataset).\\
\textbf{Prompt example:} \enquote{\emph{Grass and agricultural vegetation next to a highway, locality of Savigny-en-Revermont, Bourgogne-Franche-Comté, in the morning, during Summer.}}

\subsection{Conditional change inpainting with ControlNet}
Driven by our goal to control the new classes in the inpainted zones, we add a ControlNet (CN) architecture \cite{ControlNet} to build a model that generates realistic VHR images from a semantic segmentation map (here land cover). CN is designed to add task-specific conditioning to a pretrained image DM. It allows injecting explicit semantic guidance during the diffusion process, conditioning the outputs on some reference image (a color map of the semantic class), in addition to a text prompt.
Our training method enables CN to perform both inpainting and text-to-image synthesis, using the semantic condition. Training took 240 hours on a A100 GPU.

\subsection{Select, Mask, Change, Inpaint : the \PIPELINE~ pipeline}
\label{subsec:pipeline}

Inspired by \cite{sachdeva2023change}, the core of our method lies in our \enquote{Select and Inpaint} mechanism. 
Given a VHR image along with its semantic segmentation and instances\footnote{A panoptic segmentation is the most suitable case but not mandatory, assuming a sufficient number of instances ensures enough diversity.}, we inpaint these instances using the SD+CN inpainting method: the label of the instance is changed, the texture of the image is substituted accordingly. This results in a pair of VHR images, both with semantic maps and the corresponding semantic change maps. The complete \PIPELINE~ is provided in \cref{fig:generation_pipeline} and detailed below. It is illustrated for one image but is applied 
several times to each image of the single temporal dataset.

\vspace{-3mm}
\paragraph{\PIPELINE~uncovered.}
For an image $I_1$ with its semantic map $M_1$, and $\{O_i\}$ the set of $N_I$ instances lying on the extent of $I_1$.
\begin{compactitem}

    \item Randomly select $N_{change}$ instances among $\{O_i\}$. 
    \item For each instance, generate an \textbf{inpainting mask} : footprint+spatial buffer. 
    \item Using $M_1$, take as \enquote{$T_1$ class} the most frequent class inside the footprint: $c_1$. Then, select the larger convex hull of pixels of this class inside the buffered zone. These pixels correspond to the \textbf{change mask}: $C$.
    \item Randomly select another class $c_2$ at $T_2$ and replace the previous label by this new label.
    \item Randomly select $N_{nochange}$ random shapes in the image: such masks appear \textbf{only on the inpainting mask}, not on the change mask, letting their labels unchanged.
    \item Apply SD+CN to masked image and new semantic map.
    \item Cumulate the previous steps: we obtain our inpainting mask, change mask, and for $T_2$ a new semantic map $M_2$ and the corresponding new image $I_2$.
\end{compactitem}
{The pair of images ($I_1$,~$I_2$) with their corresponding semantic maps ($M_1$,~$M_2$), and the change map $C$ (difference between $M_2$ and $M_1$), are the samples of the hybrid dataset.

\vspace{-3mm}

\paragraph{Label and instance extraction.}
For comprehensive multi-class learning, we select FLAIR dataset \cite{garioud2023flair}, a large-scale  semantic segmentation map for 16 land cover classes over more than 800 km². FLAIR offers the advantage of covering areas where geographical instances of many FLAIR labels are publicly available\footnote{\href{https://www.data.gouv.fr/en/datasets/bd-topo-r/}{https://www.data.gouv.fr/en/datasets/bd-topo-r/}}, alleviating requiring a panoptic segmentation. Around 300,000 instance masks have been extracted for this work.

\vspace{-3mm}
\paragraph{Inpainting signature mitigation.}
Inpainting produces realistic changes but new areas tend to have sometimes a high-frequency signature, as reported by other works \cite{li2021noise,Guo_2023_CVPR}. In order to discourage the model from simply learning this inpainting noise instead of learning the actual changes between images, we also inpaint random areas of the image without changing the semantic class in a similar way to the \enquote{\textit{not changed regions}} from \cite{sachdeva2023change}. Besides, to obtain the best transition between the original image and the inpainted area, we use a continuous mask in the inference \cite{softInpainting}.

\vspace{-3mm}

\paragraph{Selection of the new label.} For each change area, we randomly select the new class $c_2$ among all classes weighted by their frequency in the whole dataset divided by their frequencies in the change area.

\vspace{-3mm}

\paragraph{Adoption of a buffer zone.} For inpainting, we add a buffer around the footprint of each modified instance. This gives SD more freedom to generate smoother borders around the changed areas. It also helps handling small spatial discrepancies between footprints and the VHR image content \cite{Gressin_2013}.

\subsection{Structure of the \FLAIRChange~dataset}

For each image in the FLAIR training set, we generate three unique modified images, resulting in 180,000 synthetic images, each with a corresponding 512$\times$512 semantic map covering 16 distinct semantic classes. We call this dataset \FLAIRCHANGE, short for FLAIR Synthetic Change. \FLAIRCHANGE~includes approximately 80 billion pixels with a ground resolution of 0.2$\:$m per pixel. Note that we can expand the dataset by pairing synthetic images that were generated (differently) from a single real image, process that we use at training time, effectively doubling the number of image pairs.
\cref{tab:synthetic_datasets} compares \FLAIRCHANGE~with other synthetic change detection datasets and shows that \FLAIRCHANGE~offers a much broader range of semantic changes beyond simple building modifications, is of higher spatial resolution, and is nearly ten times larger than comparable datasets. It should also be noted that the proportion of changes in \FLAIRCHANGE~is closer to the real scenarios observed in Europe (prevalence of 5\% change) than other academic datasets (see \cref{tab:target_datasets}), excepting HiUCD.

\paragraph{Quality assessment of \FLAIRCHANGE.} ControlNet is the most error-prone module: it may sometimes fail to respect the semantic map. We evaluated how unchanged the semantic map remains by performing 15-class semantic segmentation on real and generated images, thanks to a UNet trained for this task, leading to a satisfactory error rate below 20\% between both outputs. 
We also leveraged standard metrics for image generation evaluation to assess the realism of the synthesis.
We got an Inception Score \cite{salimans2016improved} of 6.2 on the inpainted images, which is comparable to real images, and a low FID score \cite{heusel2017gans} of 0.43 between inpainted and real images.

\section{Experiments}
\label{sec:exp}

We conduct various transfer learning experiments to validate hybrid data pretraining, comparing \FLAIRCHANGE~with the fully artificial SyntheWorld \cite{SyntheWorld}, the only available synthetic dataset with a size comparable to ours (see \cref{tab:synthetic_datasets}).

\subsection{Pretraining approaches with 
\FLAIRCHANGE} 
\label{subsec:tl_expes}
We assess \FLAIRCHANGE's effectiveness by exploiting existing open datasets as targets under various settings (see \cref{tab:target_datasets} and Supplementary Material). Here, we first present the different transfer learning approaches adopted, and then, dive into the deep models and datasets considered (\cref{subsec:model_archi}). Our objective is to demonstrate how pretraining on a hybrid dataset improves performance, for both binary and semantic change detection. \\
The four scenarios are:
\begin{compactitem} 
\item {\bf Sequential Transfer Learning.} Models are pretrained on \FLAIRCHANGE, and then fine-tuned on the target data. 
\item {\bf Low Data Regime.} The same as above but with a fraction of the available training set.
\item {\bf Mixed Training.} Models are trained on a blend of \FLAIRCHANGE~and the target data. 
\item {\bf Zero-Shot Learning.} A model pretrained on \FLAIRCHANGE~is directly applied on the target dataset, with only a remapping of \FLAIRCHANGE~classes to align labels. 
\end{compactitem}

\subsection{Experimental protocol}
\label{subsec:model_archi}
\subsubsection{Target datasets and metrics}
\label{sec:target_datasets}
Our transfer experiments span five diverse datasets, among which three contain semantic labels (\cref{tab:target_datasets}). This ensures variety in landscapes, sizes, and change types, \textit{i.e.}, a comprehensive evaluation. Among the plethora of (S)CD datasets, we selected the most popular ones: Levir-CD \cite{LevirCD}, S2Looking \cite{s2looking}, SECOND \cite{SECOND} and HiUCD-mini \cite{hiucd}.\\
We used classical metrics from the change detection literature: Intersection over Union (IoU) and F1 score for binary cases and mainly the Semantic Change Segmentation score (SCS) from \cite{Toker_2022_CVPR} for semantic evaluation. In this latter case, the Separated Kappa (SeK) from \cite{SECOND} is also informative though it tends to be more biased towards detection task than semantic prediction.\\
We added some experiments related to HiUCD-XL \cite{TIAN2022164}, a follow-up dataset of HiUCD-mini \cite{hiucd} using semantic mIoU and change mIoU, metrics provided on the evaluation platform\footnote{\href{https://www.codabench.org/competitions/3485/}{https://www.codabench.org/competitions/3485/}} as test labels are kept private.

\begin{table}[ht!]
\caption{\textbf{Target datasets.} Overview of the real-world datasets used for our transfer learning experiments.}
\label{tab:target_datasets}
\centering
\scriptsize
\renewcommand{\arraystretch}{1.5}
\setlength{\tabcolsep}{2pt}
\begin{tabular}{lrrrcrr}
\toprule
Dataset & Change ($\%$) & Pixels $\times10^6$& GSD (m) & Classes & Period & Zone\\ \midrule
Levir-CD \cite{LevirCD} & $28.4$ & 1,067 & 0.5 & 1 &  2002/18 & Texas \\ 
S2Looking \cite{s2looking} & $25.6$ & 8,389 &  0.3-0.8 & 1 & 2017/20 & World \\ 
SECOND \cite{SECOND} & $20.0$ &  1,222 & 0.3-0.6 & 6 & ? & China \\ 
HiUCD-mini \cite{hiucd} & $6.6$ &  567 & 0.1 & 9 & 2017/19 & Estonia \\ 
HiUCD-XL \cite{TIAN2022164} & $1.2$ & 10,695 & 0.1 & 9 & 2017/19 & Estonia \\ 
\midrule
\textbf{\FLAIRCHANGE~ (Ours)} & $\textbf{7.0}$ & \textbf{80,740} &  \textbf{0.2} & \textbf{16} & 2018/21 & France \\ 
\bottomrule
\end{tabular}
\end{table}

\subsubsection{Model architecture}
\label{sec:model_archi_dchan}
We consider that designing a new architecture is out of the scope of this paper and that numerous solutions have already been proposed (see \cref{sec:related}). Therefore, we initially used a simple U-Net model \cite{CDRealityCheck} with a segmentation head for semantic change detection.  Although effective for binary change, it lacked expressiveness for semantic tasks. 
Similarly to \cite{daudt2018hrscd}, we developed a simple \textbf{Dual UNet model} composed of 2 \clem{nearly identical} parallel UNets for (i) change detection, and (ii) semantic segmentation, with ResNet-50 backbones pretrained 
on ImageNet and trained in a \textit{multi-task setting}. The change branch incorporates features from the semantic encoder. 
We keep the training strategy simple and constant to focus on
data and transfer, this enhances interpretability.

\vspace{-2mm}
\paragraph{Can a binary model be sufficient ?}
We tested both a pure binary CD model and the Dual UNet, for which, when, switching to binary-only tasks, the semantic decoder is inactive but the encoding part remains active (as it provides features to the change detection branch). Such \textit{selective change detection} ability allows training on a multi-class dataset and focusing on specific binary tasks. Experiments show that even for binary tasks, the Dual UNet outperforms the binary-only model, \textit{e.g.,} achieving an IoU of 0.83 on Levir-CD (compared to 0.80). Learning more information than just change pixels, the Dual UNet gains a deeper scene understanding, leading to more reliable predictions.

\paragraph{Existing architectures.}
We tested more complex transformer-based architectures, namely ChangeFormer \cite{ChangeFormer}, MambaCD \cite{ChangeMamba}, and SCanNet \cite{SCanNet}, both on HiUCD-mini and SECOND datasets. SCanNet performed best, balancing high performance and straightforward implementation due to its lightweight design (see Supplementary Material). SCanNet is kept hereafter for comparisons \clem{(see \cref{tab:sequential_scd})}.

\paragraph{Transfer for varying semantic classes.}
    In the semantic case, we faced the issue of distinct classes between  source and target datasets. This required to choose which classes should be kept for model training.
    \begin{compactitem}
        \item \textbf{Sequential case:} for both pretraining and fine-tuning strategies, we use the classes of the processed dataset. The semantic head of the model is reinitialized and adapted to the new number of classes before fine-tuning.
        \item \textbf{Mixed case:} Classes from the target dataset are used, manually mapping  pretraining classes to their closest counterparts.
    \end{compactitem}

\subsubsection{Experimental setting}
We apply basic augmentations on the images: horizontal and vertical flips, 90° rotations, and small rescaled crops (scales above 0.8). Normalization by mean and variance from the target dataset is used in all experiments. 
For optimization, we use the AdamW algorithm with default parameters (0.01 weight decay), and an initial learning rate of 3e-4 with exponential decay (0.99 per epoch). 
For sequential training, models are pretrained for 40h on a A100 GPU, with fine-tuning requiring up to 20h on a V100. For mixed training, we limit the total training time to 20h on a A100.

\subsection{Results and Analysis}

\subsubsection{Sequential training}

\paragraph{Binary Case.}
Models pretrained on synthetic datasets outperform those without pretraining on all target datasets (\cref{fig:comp_transfer_datasets}).
SyntheWorld maintains commendable performance on the the single-class \enquote{\textit{building}} change case, similarly to Levir-CD. 
Pretraining on \FLAIRCHANGE~yields the strongest results, compared to SyntheWorld on all five target datasets, with gains of 2-3\% of IoU.
Details are provided in the Supplementary Material.

\begin{figure}[!ht]
\centering

\begin{subfigure}{0.495\columnwidth} 
    \centering
    \begin{tikzpicture}
      \begin{axis}[
       title={\parbox{\linewidth}{\centering (a) Levir-CD\label{subfig:comp_LevirCD}}},
        width=0.9\textwidth,
        height=0.9\textwidth,
        ybar,
        bar width=20pt,
        enlarge x limits={abs=10pt},
        ylabel={IoU},
        symbolic x coords={From Scratch, SyntheWorld, FlairChange},
        xtick=\empty, %
        y tick label style={font=\footnotesize},
        ymin=0.82, ymax=0.86,
        bar shift=0pt,
      ]
        \addplot[draw=none,fill=baselinecolor] coordinates {(From Scratch,0.832)};
        \addplot[draw=none,fill=synthecolor] coordinates {(SyntheWorld,0.836)};
        \addplot[draw=none,fill=flaircolor] coordinates {(FlairChange,0.838)};
      \end{axis}
    \end{tikzpicture}
\end{subfigure}
\begin{subfigure}{0.495\columnwidth}  
    \centering 
    \begin{tikzpicture}
      \begin{axis}[
        title={\parbox{\linewidth}{\centering (b) S2Looking\label{subfig:comp_S2}}},
        width=0.9\textwidth,
        height=0.9\textwidth,
        ybar,
        bar width=20pt,
        enlarge x limits={abs=10pt},
        symbolic x coords={From Scratch, SyntheWorld, FlairChange},
        xtick=\empty, %
        y tick label style={font=\footnotesize},
        ymin=0.42, ymax=0.48,
        bar shift=0pt,
      ]
        \addplot[draw=none,fill=baselinecolor] coordinates {(From Scratch,0.442)};
        \addplot[draw=none,fill=synthecolor] coordinates {(SyntheWorld,0.456)};
        \addplot[draw=none,fill=flaircolor] coordinates {(FlairChange,0.464)};
      \end{axis}
    \end{tikzpicture}
\end{subfigure}
\begin{subfigure}[t]{0.495\columnwidth} 
    \centering 
    \begin{tikzpicture}
      \begin{axis}[
        title={\parbox{\linewidth}{\centering (c) SECOND\label{subfig:comp_SECOND}}},
        width=0.9\textwidth,
        height=0.9\textwidth,
        ybar,
        bar width=20pt,
        enlarge x limits={abs=10pt},
        ylabel={IoU},
        symbolic x coords={From Scratch, SyntheWorld, FlairChange},
        xtick=\empty, %
        y tick label style={font=\footnotesize},
        ymin=0.52, ymax=0.56,
        bar shift=0pt,
      ]
        \addplot[draw=none,fill=baselinecolor] coordinates {(From Scratch,0.533)};
        \addplot[draw=none,fill=synthecolor] coordinates {(SyntheWorld,0.544)};
        \addplot[draw=none,fill=flaircolor] coordinates {(FlairChange,0.545)};
      \end{axis}
    \end{tikzpicture}
\end{subfigure}
\begin{subfigure}[t]{0.495\columnwidth}  
    \centering 
    \begin{tikzpicture}
      \begin{axis}[
        title={\parbox{\linewidth}{\centering (d) HiUCD-mini\label{subfig:comp_HiUCD_mini}}},
        width=0.9\textwidth,
        height=0.9\textwidth,
        ybar,
        bar width=20pt,
        enlarge x limits={abs=10pt},
        symbolic x coords={From Scratch, SyntheWorld, FlairChange},
        xtick=\empty, %
        y tick label style={font=\footnotesize},
        ymin=0.56, ymax=0.64,
        bar shift=0pt,
      ]
        \addplot[draw=none,fill=baselinecolor] coordinates {(From Scratch,0.606)};
        \addplot[draw=none,fill=synthecolor] coordinates {(SyntheWorld,0.580)};
        \addplot[draw=none,fill=flaircolor] coordinates {(FlairChange,0.630)};
      \end{axis}
    \end{tikzpicture}
\end{subfigure}
\begin{tikzpicture}
\begin{axis}[
    hide axis,
    xmin=0, xmax=1, ymin=0, ymax=1,
    height=50px,
    legend columns=3,
    legend style={at={(0,0)}, anchor=north, font=\small, /tikz/every even column/.append style={column sep=10pt}}
]
\addlegendimage{fill=baselinecolor,legend image code/.code={\draw[fill=baselinecolor] (0,-4px) rectangle (20px,5px);}}   
\addlegendentry{Baseline}
\addlegendimage{fill=synthecolor,legend image code/.code={\draw[fill=synthecolor] (0,-4px) rectangle (20px,5px);}}   
\addlegendentry{SyntheWorld}
\addlegendimage{fill=flaircolor,legend image code/.code={\draw[fill=flaircolor] (0,-4px) rectangle (20px,5px);}}  
\addlegendentry{\FLAIRCHANGE}
\end{axis}
\end{tikzpicture}

    \caption[  ]{ \textbf{Sequential training in binary change detection.} After pretraining our Dual U-Net on either {\color{baselinecolor}{nothing}}, {\color{synthecolor}{SyntheWorld}} or {\color{flaircolor}{\FLAIRCHANGE}}, we finetune and test it on each of the 4 target datasets and represent the binary IoU.}
    \label{fig:comp_transfer_datasets}
\end{figure}
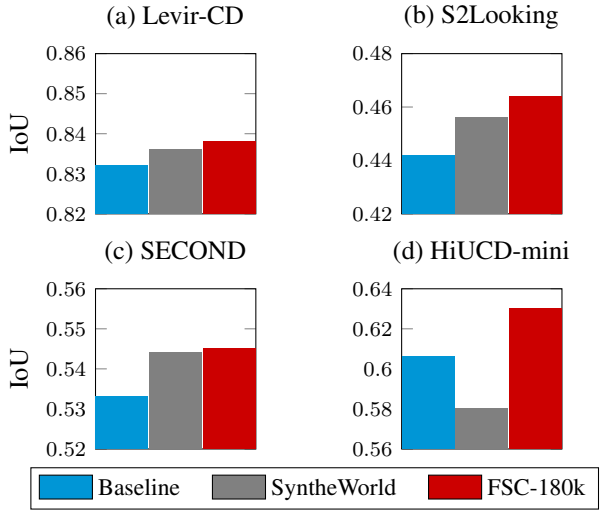

\paragraph{Semantic Case.} Pretraining on \FLAIRCHANGE~outperforms the baseline and SyntheWorld on the three SCD datasets in the sequential \clem{mode (\cref{tab:sequential_scd}).}
Indeed, pretraining on SyntheWorld, that focuses solely on the \enquote{\textit{building}} class, shows limitations in multiclass tasks on HiUCD-mini and HiUCD-XL compared to the baseline.
Our \FLAIRCHANGE~dataset is particularly beneficial for HiUCD-mini and HiUCD-XL. The types of landscapes and spatial resolution are relatively similar, which facilitates effective transfer learning. For HiUCD-mini, our pretraining results in a SeK of 0.19 and a SCS of 0.78, representing respective increases of 11\% and 7\% compared to the baseline. The gains rise to 28\% and 8\% against SyntheWorld.
For SECOND, SyntheWorld partially narrows the gap: SeK of 0.17 and a SCS of 0.84, while the \FLAIRCHANGE's pretrained Dual UNet achieves 0.18 and 0.87. 

\begin{table}[ht]
    \caption{\textbf{Sequential training in the semantic change detection framework.} We report the binary change detection scores (IoU or Chg. mIoU) as well as semantic scores (SCS or mIoU) on three datasets in the sequential scheme. Models were initially pretrained ({\color{synthecolor}{SyntheWorld}} or {\color{flaircolor}{\FLAIRCHANGE}}) or not ({\color{baselinecolor}{Baseline}}).}
    \label{tab:sequential_scd}
\small
    \begin{tabular}{llcccc}
    \toprule
           & & \multicolumn{2}{c}{SECOND} & \multicolumn{2}{c}{HiUCD-mini}  \\
           \cmidrule(l){3-4} \cmidrule(l){5-6}
         Pretraining & Model & IoU & SCS & IoU & SCS  \\
         \midrule
         \rowcolor{black!5} {\color{baselinecolor} None} &Dual UNet& 0.53 & 0.83 & 0.61 & 0.73 \\
         {\color{synthecolor}SyntheWorld} & Dual UNet & 0.54 & 0.84 & 0.58 & 0.73 \\
      \rowcolor{black!5}   {\color{flaircolor}\FLAIRCHANGE} &Dual UNet& 0.55 & \textbf{0.89} & \textbf{0.63} & \textbf{0.78}  \\
         \midrule
          {\color{baselinecolor} None} & SCanNet & 0.54 & 0.84& 0.60 & 0.66  \\
       \rowcolor{black!5}   {\color{flaircolor}\FLAIRCHANGE} & SCanNet & \textbf{0.56} & \textbf{0.89} & \textbf{0.63} & 0.71 \\
        \bottomrule
    \end{tabular}
    
    \vspace{5px}
    \centering
    \begin{tabular}{lcc}
        \toprule
           & \multicolumn{2}{c}{HiUCD-XL} \\
           \cmidrule(l){2-3}
         Pretraining  & Chg. mIoU & mIoU \\
         \midrule
\rowcolor{black!5}          {\color{baselinecolor} None}   & 0.17 &  0.58 \\
         {\color{synthecolor}SyntheWorld}  & 0.17 &  0.58 \\
\rowcolor{black!5}          {\color{flaircolor}\FLAIRCHANGE}  & \textbf{0.19} &  \textbf{0.60} \\
        \bottomrule
    \end{tabular}
\end{table}

\paragraph{Dependence on a CD model.}
We compared two CD model by running the same experiments on SECOND and HiUCD-mini with a SCanNet model (\cref{tab:sequential_scd}). 
We were unable to reproduce the results reported in the original paper \cite{SCanNet}, despite following the described learning scheme. SCanNet still performs well on the SECOND dataset, achieving a SCS of 0.84, compared to 0.83 for the Dual UNet.
The opposite trend is observed on HiUCD, where the Dual UNet reaches a SCS of 0.73, while SCanNet reaches 0.66. 
As neither of the two models stands out, we mainly present the results obtained with our Dual UNet.

\subsubsection{Mixed training}
Our experiments show, that the optimal ratio between synthetic and target data lies around 50\% for HiUCD-XL, S2Looking or SECOND (\cref{fig:Mixed_five_datasets}) conversely to the conclusions in \cite{SyntheWorld}, where performance always increases with the amount of target data included in the training mix. Nevertheless, the optimal ratio is 90\% for smaller dataset such as Levir-CD or HiUCD-mini, that may be too specific. However, the mixed training strategy proves to be really effective, outperforming the sequential one on HiUCD-XL with a balanced blend as training base, as noticed in \cref{fig:Mixed_five_datasets}, compared to \cref{tab:sequential_scd}.
We can deduce that:
\begin{compactitem}
\item \textbf{Sequential training maybe suboptimal}: The lengthy fine-tuning process can lead the model to \enquote{forget} pretraining, and instead overfits on the target data, which finally reduces the benefits of pretraining.
\item In contrast, \textbf{mixed training} augments the initial set and \textbf{adds diversity} throughout the training process. The model sees as many images from the target dataset as in the sequential mode, but continues to benefit from the pretraining dataset.
\item A \enquote{\textbf{balanced blend}} of synthetic and target data appears optimal, \textbf{providing a sufficient frequency of target images} for exploitation while strategically incorporating new synthetic images to enable exploratory learning and notably avoid overfitting.
\end{compactitem}

\begin{figure*}
\small
\begin{subfigure}[t]{0.400\columnwidth}
    \centering
    \begin{tikzpicture}
    \begin{axis}[
        title={\parbox{\textwidth}{\centering (a) Levir-CD\label{subfig:levir_Mixed}}},
        xlabel={Mixed Ratio (\%)},
        ylabel={\small IoU},
        xtick={0, 20, 50, 90, 100},
        xticklabels={0, 20, 50, \hspace{-5pt}90, \hspace{10pt}100},
        ymin=0.79, ymax=0.91,
        width=1.1\columnwidth,
        height=\columnwidth, 
        legend style={font=\small},
        legend image post style={scale=0.7},
        grid=both,
    ]
    \addplot[color=synthecolor, thick, mark=square*] coordinates {(20, 0.81) (50, 0.833) (90, 0.838) (100, 0.832)}; 
    \addplot[color=flaircolor, thick, mark=triangle*] coordinates {(20, 0.817) (50, 0.832) (90, 0.844) (100, 0.832)};
    \end{axis}
    \end{tikzpicture}
\end{subfigure}
\hspace{-0.1cm}
\begin{subfigure}[t]{0.400\columnwidth} 
    \centering
    \begin{tikzpicture}
    \begin{axis}[
        title={\parbox{\linewidth}{\centering (b) S2Looking\label{subfig:S2_Mixed}}},
        xlabel={Mixed Ratio (\%)},
        xtick={0, 20, 50, 90, 100},
        xticklabels={0, 20, 50, \hspace{-5pt}90, \hspace{10pt}100},
        ymin=0.39, ymax=0.51,
        width=1.1\columnwidth,
        height=\columnwidth, 
        legend style={font=\small},
        legend image post style={scale=0.7},
        grid=both,
    ]
    
   \addplot[color=synthecolor, thick, mark=square*] coordinates {(20, 0.428) (50, 0.476) (90, 0.477) (100, 0.442)}; 
    \addplot[color=flaircolor, thick, mark=triangle*] coordinates {(20, 0.441) (50, 0.488) (90, 0.486) (100, 0.442)};
    
    \end{axis}
    \end{tikzpicture}
\end{subfigure}
\hspace{-0.3cm}
\begin{subfigure}[t]{0.400\columnwidth}
    \centering
    \begin{tikzpicture}
    \begin{axis}[
        title={\parbox{\linewidth}{\centering (c) SECOND\label{subfig:second_Mixed}}},
        xlabel={Mixed Ratio (\%)},
        ylabel={\small SCS},
        xtick={0, 20, 50, 90, 100},
        xticklabels={0, 20, 50, \hspace{-5pt}90, \hspace{10pt}100},
        ymin=0.79, ymax=0.91,
        width=1.1\columnwidth,
        height=\columnwidth, 
        legend style={font=\small},
        legend image post style={scale=0.7},
        grid=both,
    ]
    
    \addplot[color=synthecolor, thick, mark=square*] coordinates {(20,0.828) (50,0.841) (90,0.846) (100,0.830)};
    \addplot[color=flaircolor, thick, mark=triangle*] coordinates {(20,0.878) (50,0.889) (90,0.885) (100,0.830)};
    
    \end{axis}
    \end{tikzpicture}
\end{subfigure}
\hspace{-0.1cm}
\begin{subfigure}[t]{0.400\columnwidth} %
    \centering
    \begin{tikzpicture}
    \begin{axis}[
        title={\parbox{\linewidth}{\centering (d) HiUCD-mini\label{subfig:mini_Mixed}}},
        xlabel={Mixed Ratio (\%)},
        xtick={0, 20, 50, 90, 100},
        xticklabels={0, 20, 50, \hspace{-5pt}90, \hspace{10pt}100},
        ymin=0.65, ymax=0.81,
        width=1.1\columnwidth,
        height=\columnwidth, 
        legend style={font=\small},
        legend image post style={scale=0.7},
        grid=both,
    ]
    \addplot[color=synthecolor, thick, mark=square*] coordinates {(20,0.680) (50,0.721) (90,0.713) (100,0.73)};

    \addplot[color=flaircolor, thick, mark=triangle*] coordinates {(20,0.736) (50,0.761) (90,0.783) (100,0.73)}; %
    
    \end{axis}
    \end{tikzpicture}
\end{subfigure}
\hspace{-0.5cm}
\begin{subfigure}[t]{0.400\columnwidth} %
    \centering
    \begin{tikzpicture}
    \begin{axis}[
        title={\parbox{\linewidth}{\centering (e) HiUCD-XL\label{subfig:XL_Mixed}}},
        xlabel={Mixed Ratio (\%)},
        ylabel={\small Chg mIoU},
        xtick={0, 20, 50, 90, 100},
        xticklabels={0, 20, 50, \hspace{-5pt}90, \hspace{10pt}100},
        ymin=0.09, ymax=0.21,
        width=1.1\columnwidth,
        height=\columnwidth, 
        legend style={font=\small},
        legend image post style={scale=0.7},
        grid=both,
    ]
    
\addplot[color=synthecolor, thick, mark=square*] coordinates {(20, 0.101) (50, 0.186) (90, 0.173)  (100, 0.172)};%
\addplot[color=flaircolor, thick, mark=triangle*] coordinates {(20, 0.154) (50, 0.195) (90, 0.181) (100, 0.172)};%

    \end{axis}
    \end{tikzpicture}
\end{subfigure}

\centering
\begin{tikzpicture}
\begin{axis}[
    hide axis,
    height=50px,
    xmin=0, xmax=1, ymin=0, ymax=1,
    legend columns=3, %
    legend style={at={(0.5,1.05)}, anchor=south, font=\small, /tikz/every even column/.append style={column sep=10pt}}
]
Mix Ratio (\%)
\addlegendimage{color=synthecolor, thick, mark=square*}
\addlegendentry{SyntheWorld}
\addlegendimage{color=flaircolor, thick, mark=triangle*}
\addlegendentry{\FLAIRCHANGE}
\end{axis}
\end{tikzpicture}

\caption[  ]{\textbf{Mixed training (Binary and Semantic).}
    {\small Training on a blend of target and synthetic/hybrid ({\color{synthecolor}SyntheWorld} or {\color{flaircolor}\FLAIRCHANGE}) train sets, containing a ratio of x\% samples from the target (including repetitions). Testing is performed on the target test set. 100\% corresponds to fine-tuning exclusively on target dataset (without pretraining).} }%
\label{fig:Mixed_five_datasets}
\end{figure*}
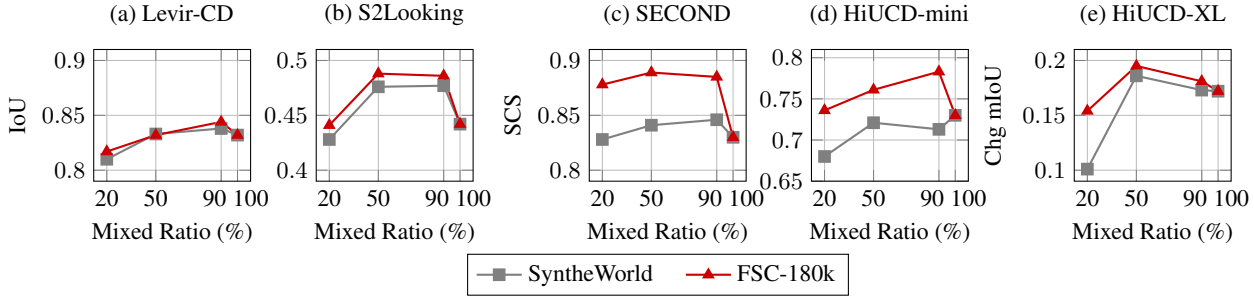

\subsubsection{Low Data regime}

We apply a sequential scheme but limit fine-tuning to small subsets of the target data: specifically, 1\%, 10\%, and 30\% of the training set. Each experiment was repeated 10 times, with a random sampling of the training set. 
We consider these to be the most meaningful experiments, as this scenario closely matches real-world use cases. 
The primary objective of pretraining is to develop a model that can either \textbf{swiftly} adapt to new target datasets or perform well when \textbf{only limited training data is available}.
One can observe on \cref{fig:LowDataSemantic} a clear advantage for pretrained models at very low level (1\%). With such little data, the model cannot fully learn the specificities of the target dataset and relies primarily on its general knowledge and on the task itself (\textit{detecting changes}), rather than on dataset-specific features. Unsurprisingly, models pretrained for this task perform significantly better.
However, as the amount of available data increases, the gap tends to narrow. Models increasingly rely on data itself, to which they have equal access, yet pretraining on \FLAIRCHANGE~still yields better performance than using SyntheWorld or training from scratch. 

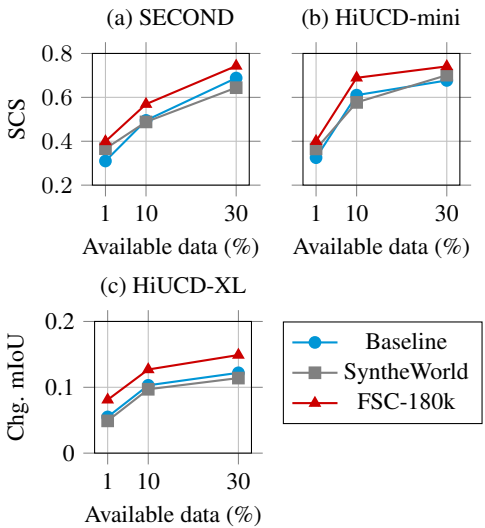
\begin{figure}[ht!]
\small
\centering

\begin{subfigure}[b]{0.400\columnwidth}
    \centering
    \begin{tikzpicture}
    \begin{axis}[
         title={\parbox{\linewidth}{\centering (a) SECOND\label{subfig:SECOND_lowdata}}},
        xlabel={Available data (\%)},
        ylabel={SCS},
        xtick={1, 10, 30},
        xticklabels={1, 10, 30},
        ymin=0.2, ymax=0.8,
        width=1.1\columnwidth,
        height=\columnwidth, 
        tick align=outside,
        legend style={font=\small},
        grid=both,
    ]
    
    \addplot[color=baselinecolor, thick, mark=*] coordinates {(1,0.310) (10,0.495) (30,0.688)};

    \addplot[color=synthecolor, thick, mark=square*] coordinates {(1,0.366) (10,0.488) (30,0.644)};

    \addplot[color=flaircolor, thick, mark=triangle*] coordinates {(1,0.399) (10,0.569) (30,0.743)};
    
    \end{axis}
    \end{tikzpicture}
\end{subfigure}
\hspace{-0.2cm}
\begin{subfigure}[b]{0.400\columnwidth} 
        \centering
    \begin{tikzpicture}
    \begin{axis}[
        title={\parbox{\linewidth}{\centering (b) HiUCD-mini\label{subfig:HiUCDmini_lowdata}}},
        xlabel={Available data (\%)},
        xtick={1, 10, 30},
        xticklabels={1, 10, 30},
        ymin=0, ymax=0.6,
        width=1.1\columnwidth, 
        height=\columnwidth, 
        yticklabels={},
        tick align=outside,
        grid=both,
    ]
    
    \addplot[color=baselinecolor, thick, mark=*] coordinates {(1,0.125) (10,0.410) (30,0.477)};
    \addplot[color=synthecolor, thick, mark=square*] coordinates {(1,0.165) (10,0.377) (30,0.501)};
    \addplot[color=flaircolor, thick, mark=triangle*] coordinates {(1,0.200) (10,0.489) (30,0.541)};
    
    \end{axis}
    \end{tikzpicture}
\end{subfigure} \\
\hspace{-3.4cm}
\begin{subfigure}[b]{0.400\columnwidth}
    \centering
    \begin{tikzpicture}
    \begin{axis}[
         title={\parbox{\linewidth}{\centering (c) HiUCD-XL\label{subfig:hiucdxl_lowdata}}},
        xlabel={Available data (\%)},
        ylabel={Chg. mIoU},
        xtick={1, 10, 30},
        xticklabels={1, 10, 30},
        ytick={0, 0.1, 0.2},
        yticklabels={0, 0.1, 0.2},
        ymin=0, ymax=0.2,
        width=1.1\columnwidth,
        height=\columnwidth, 
        tick align=outside,
        grid=both,
        legend style={font=\small},
     legend columns=1, %
    legend style={at={(1.2,1)},anchor=north west, font=\small, /tikz/every even column/.append style={column sep=10pt}}
    ]
    \addlegendimage{color=baselinecolor, thick, mark=*}
    \addlegendentry{Baseline}
    \addlegendimage{color=synthecolor, thick, mark=square*}
    \addlegendentry{SyntheWorld}
    \addlegendimage{color=flaircolor, thick, mark=triangle*}
    \addlegendentry{\FLAIRCHANGE}
    
    \addplot[color=baselinecolor, thick, mark=*] coordinates {(1,0.055) (10,0.103) (30,0.122)};

    \addplot[color=synthecolor, thick, mark=square*] coordinates {(1,0.049) (10,0.097) (30,0.114)};

    \addplot[color=flaircolor, thick, mark=triangle*] coordinates {(1,0.081) (10,0.127) (30,0.149)};
    
    \end{axis}
    \end{tikzpicture}
    \end{subfigure}

\caption[  ]{\textbf{Low data regime (Semantic).}
    {\small Fine-tuning on a limited part of the target train set (either 1\%, 10\% or 30\% of randomly sampled examples) and evaluating on the whole test set. Models are initially pretrained ({\color{synthecolor}{SyntheWorld}} or {\color{flaircolor}{\FLAIRCHANGE}}) or not ({\color{baselinecolor}{Baseline}}). Metrics are averaged on 10 runs for SECOND and HiUCD-mini.}}
\label{fig:LowDataSemantic}
\end{figure}

\subsubsection{Zero-Shot evaluation}
While models pretrained on SyntheWorld struggle to transfer effectively without target sample exposure, pretraining on \FLAIRCHANGE~delivers respectable results, achieving IoU scores of 0.37 on SECOND and 0.36 on HiUCD-mini, showing performance drops of 31\% and 40\%, compared to the fully supervised models (details are provided in the Supplementary Material). However, \FLAIRCHANGE~is less effective on Levir-CD, where the performance drop reaches 60\% (0.33 vs. 0.83 IoU). We argue that performance in this scenario is directly linked to dataset similarities: the limited realism in SyntheWorld hinders transferability, while \FLAIRCHANGE’s diverse landscapes and sparse changes further limit its effectiveness. For the semantic task, pretraining on \FLAIRCHANGE~enhances the model's understanding of change trajectories, achieving SCS of 0.24 on SECOND and 0.25 on HiUCD-mini.

\subsubsection{Qualitative assessment}

Qualitative analysis of the various segmentation results indicates that \FLAIRCHANGE~pretraining significantly enhances the network's ability to localize features in input images. This improvement leads to both more accurate segmentation within VHR images and higher semantic prediction accuracy for main classes of interest in the semantic case (\textit{e.g.}, \textit{buildings}, \textit{roads}, \textit{bare land} for HiUCD-mini, \textit{buildings} or \textit{non-vegetated ground surface} for SECOND). Qualitative examples for the different datasets are provided in the Supplementary Material.

\section{Conclusion}

We introduced \PIPELINE, a new, simple bi-temporal change detection generation pipeline. It lies on Stable Diffusion and ControlNet for creating hybrid content of VHR images. \PIPELINE~leverages an existing large-scale semantic land cover map to overcome the lack of diversity and size of real-world datasets. We provide \FLAIRCHANGE~, a pretraining dataset based on FLAIR \cite{garioud2023flair}, which includes bi-temporal real/inpainted images, and their associated semantic maps. 
The proposed solution was evaluated for transferability, on five datasets over four distinct training setups. Our experiments on both binary and semantic cases demonstrated strong transferability, confirming \FLAIRCHANGE~robustness and adaptability as well as \PIPELINE~relevance. 

\FloatBarrier
\pagebreak

\section*{Acknowledgements} 
This work was granted access to the HPC resources of IDRIS under the allocations AD011014690 and AD011014286 made by GENCI.
We thank Ruben Gres for inspiring discussions and valuable feedback about ControlNet and Stable Diffusion.

{\small
\bibliographystyle{ieee_fullname}
\bibliography{egbib}

\begin{thebibliography}{10}\itemsep=-1pt

\bibitem{softInpainting}
Andrew.
\newblock How to use {{Soft Inpainting}} - {{Stable Diffusion Art}}.
\newblock \url{https://stable-diffusion-art.com/soft-inpainting/}, 2024.

\bibitem{ChangeFormer}
Wele Gedara~Chaminda Bandara and Vishal~M. Patel.
\newblock A transformer-based siamese network for change detection.
\newblock In {\em IGARSS}, pages 207--210, 2022.

\bibitem{Burke_Science}
Marshall Burke, Anne Driscoll, David~B. Lobell, and Stefano Ermon.
\newblock Using satellite imagery to understand and promote sustainable development.
\newblock {\em Science}, 371(6535), 2021.

\bibitem{CAO2023113371}
Yinxia Cao and Xin Huang.
\newblock A full-level fused cross-task transfer learning method for building change detection using noise-robust pretrained networks on crowdsourced labels.
\newblock {\em Remote Sensing of Environment}, 284:113371, 2023.

\bibitem{chen2021adversarial}
Hao Chen, Wenyuan Li, and Zhenwei Shi.
\newblock Adversarial instance augmentation for building change detection in remote sensing images.
\newblock {\em IEEE Transactions on Geoscience and Remote Sensing}, 60:1--16, 2021.

\bibitem{LevirCD}
Hao Chen and Zhenwei Shi.
\newblock A spatial-temporal attention-based method and a new dataset for remote sensing image change detection.
\newblock {\em Remote Sensing}, 12(10), 2020.

\bibitem{ChangeMamba}
Hongruixuan Chen, Jian Song, Chengxi Han, Junshi Xia, and Naoto Yokoya.
\newblock Changemamba: Remote sensing change detection with spatiotemporal state space model.
\newblock {\em IEEE Transactions on Geoscience and Remote Sensing}, 62:1--20, 2024.

\bibitem{CDRealityCheck}
Isaac Corley, Caleb Robinson, and Anthony Ortiz.
\newblock A change detection reality check.
\newblock {\em arXiv preprint arXiv:2402.06994}, 2024.

\bibitem{czerkawski2024exploring}
Mikolaj Czerkawski and Christos Tachtatzis.
\newblock Exploring the capability of text-to-image diffusion models with structural edge guidance for multi-spectral satellite image inpainting.
\newblock {\em IEEE Geoscience and Remote Sensing Letters}, 2024.

\bibitem{czerkawski2022deep}
Mikolaj Czerkawski, Priti Upadhyay, Christopher Davison, Astrid Werkmeister, Javier Cardona, Robert Atkinson, Craig Michie, Ivan Andonovic, Malcolm Macdonald, and Christos Tachtatzis.
\newblock Deep internal learning for inpainting of cloud-affected regions in satellite imagery.
\newblock {\em Remote Sensing}, 14(6):1342, 2022.

\bibitem{daudt2018hrscd}
{Rodrigo Caye} Daudt, Bertrand {Le Saux}, Alexandre Boulch, and Yann Gousseau.
\newblock Multitask learning for large-scale semantic change detection.
\newblock {\em Computer Vision and Image Understanding}, 187:102783, 2019.

\bibitem{daudt2018fully}
Rodrigo~Caye Daudt, Bertrand Le~Saux, and Alexandre Boulch.
\newblock Fully convolutional siamese networks for change detection.
\newblock In {\em ICIP}, pages 4063--4067, 2018.

\bibitem{SCanNet}
Lei Ding, Jing Zhang, Haitao Guo, Kai Zhang, Bing Liu, and Lorenzo Bruzzone.
\newblock Joint spatio-temporal modeling for semantic change detection in remote sensing images.
\newblock {\em IEEE Transactions on Geoscience and Remote Sensing}, 62:1--14, 2024.

\bibitem{GenerateYourOwnScotland}
Miguel Espinosa and Elliot~J. Crowley.
\newblock Generate your own scotland: Satellite image generation conditioned on maps.
\newblock {\em NeurIPS 2023 Workshop on Diffusion Models}, 2023.

\bibitem{TamingTansformers}
Patrick Esser, Robin Rombach, and Bjorn Ommer.
\newblock Taming transformers for high-resolution image synthesis.
\newblock In {\em CVPR}, pages 12873--12883, 2021.

\bibitem{fan2011pixel}
Weiliang Fan, Jing~M Chen, and Weimin Ju.
\newblock A pixel missing patch inpainting method for remote sensing image.
\newblock In {\em International Conference on Geoinformatics}, 2011.

\bibitem{SMARS}
Mario {Fuentes Reyes}, Yuxing Xie, Xiangtian Yuan, Pablo d’Angelo, Franz Kurz, Daniele Cerra, and Jiaojiao Tian.
\newblock A 2{D/3D} multimodal data simulation approach with applications on urban semantic segmentation, building extraction and change detection.
\newblock {\em ISPRS Journal of Photogrammetry and Remote Sensing}, 205:74--97, 2023.

\bibitem{garioud2023flair}
Anatol Garioud, Nicolas Gonthier, Loic Landrieu, Apolline~De Wit, Marion Valette, Marc Poup{\'e}e, Sebastien Giordano, and Boris Wattrelos.
\newblock {FLAIR} : a country-scale land cover semantic segmentation dataset from multi-source optical imagery.
\newblock In {\em NeurIPS Datasets and Benchmarks Track}, 2023.

\bibitem{Gressin_2013}
Adrien Gressin, Nicole Vincent, Clément Mallet, and Nicolas Paparoditis.
\newblock Semantic approach in image change detection.
\newblock In {\em ACIVS}, 2013.

\bibitem{Guo_2023_CVPR}
Xiao Guo, Xiaohong Liu, Zhiyuan Ren, Steven Grosz, Iacopo Masi, and Xiaoming Liu.
\newblock Hierarchical fine-grained image forgery detection and localization.
\newblock In {\em CVPR}, 2023.

\bibitem{heusel2017gans}
Martin Heusel, Hubert Ramsauer, Thomas Unterthiner, Bernhard Nessler, and Sepp Hochreiter.
\newblock Gans trained by a two time-scale update rule converge to a local nash equilibrium.
\newblock {\em Advances in neural information processing systems}, 30, 2017.

\bibitem{howarth1981procedures}
Philip~J Howarth and Gregory~M Wickware.
\newblock Procedures for change detection using landsat digital data.
\newblock {\em International Journal of Remote Sensing}, 2(3):277--291, 1981.

\bibitem{huang2022image}
Wenli Huang, Ye Deng, Siqi Hui, and Jinjun Wang.
\newblock Image inpainting with bilateral convolution.
\newblock {\em Remote Sensing}, 14(23):6140, 2022.

\bibitem{isola2017image}
Phillip Isola, Jun-Yan Zhu, Tinghui Zhou, and Alexei~A Efros.
\newblock Image-to-image translation with conditional adversarial networks.
\newblock In {\em IEEE Conference on Computer Vision and Pattern Recognition (CVPR)}, pages 1125--1134, 2017.

\bibitem{ji2018fully_WHU}
Shunping Ji, Shiqing Wei, and Meng Lu.
\newblock Fully convolutional networks for multisource building extraction from an open aerial and satellite imagery data set.
\newblock {\em IEEE Transactions on Geoscience and Remote Sensing}, 57(1):574--586, 2018.

\bibitem{khanna2023diffusionsat}
Samar Khanna, Patrick Liu, Linqi Zhou, Chenlin Meng, Robin Rombach, Marshall Burke, David~B Lobell, and Stefano Ermon.
\newblock Diffusionsat: A generative foundation model for satellite imagery.
\newblock In {\em ICLR}, 2023.

\bibitem{kolos2019procedural}
Maria Kolos, Anton Marin, Alexey Artemov, and Evgeny Burnaev.
\newblock Procedural synthesis of remote sensing images for robust change detection with neural networks.
\newblock In {\em Advances in Neural Networks}, 2019.

\bibitem{kuznetsov2020remote}
Andrey Kuznetsov and Mikhail Gashnikov.
\newblock Remote sensing image inpainting with generative adversarial networks.
\newblock In {\em ISDFS}, 2020.

\bibitem{li2021noise}
Ang Li, Qiuhong Ke, Xingjun Ma, Haiqin Weng, Zhiyuan Zong, Feng Xue, and Rui Zhang.
\newblock Noise doesn't lie: towards universal detection of deep inpainting.
\newblock In {\em IJCAI}, 2021.

\bibitem{Li_2024_CVPR}
Zhuohong Li, Wei He, Jiepan Li, Fangxiao Lu, and Hongyan Zhang.
\newblock Learning without exact guidance: Updating large-scale high-resolution land cover maps from low-resolution historical labels.
\newblock In {\em CVPR}, 2024.

\bibitem{focal}
Tsung-Yi Lin, Priya Goyal, Ross Girshick, Kaiming He, and Piotr Dollár.
\newblock Focal loss for dense object detection.
\newblock {\em IEEE Transactions on Pattern Analysis and Machine Intelligence}, 42(2):318--327, 2020.

\bibitem{lv2021land}
Zhiyong Lv, Tongfei Liu, J{\'o}n~Atli Benediktsson, and Nicola Falco.
\newblock Land cover change detection techniques: Very-high-resolution optical images: A review.
\newblock {\em IEEE Geoscience and Remote Sensing Magazine}, 10(1):44--63, 2021.

\bibitem{ma2024transfer}
Yuchi Ma, Shuo Chen, Stefano Ermon, and David~B Lobell.
\newblock Transfer learning in environmental remote sensing.
\newblock {\em Remote Sensing of Environment}, 301:113924, 2024.

\bibitem{marin2021sssgan}
Javier Mar{\'\i}n and Sergio Escalera.
\newblock Sssgan: Satellite style and structure generative adversarial networks.
\newblock {\em Remote Sensing}, 13(19):3984, 2021.

\bibitem{meng2017sparse}
Fan Meng, Xiaomei Yang, Chenghu Zhou, and Zhi Li.
\newblock A sparse dictionary learning-based adaptive patch inpainting method for thick clouds removal from high-spatial resolution remote sensing imagery.
\newblock {\em Sensors}, 17(9):2130, 2017.

\bibitem{ijgi12110450}
Ali Mirzaei, Hossein Bagheri, and Iman Khosravi.
\newblock Enhancing crop classification accuracy through synthetic {SAR-O}ptical data generation using deep learning.
\newblock {\em ISPRS International Journal of Geo-Information}, 12(11), 2023.

\bibitem{nemmour2006multiple}
Hassiba Nemmour and Youcef Chibani.
\newblock Multiple support vector machines for land cover change detection: An application for mapping urban extensions.
\newblock {\em ISPRS Journal of Photogrammetry and Remote Sensing}, 61(2):125--133, 2006.

\bibitem{OpenStreetMap}
{OpenStreetMap contributors}.
\newblock {Planet dump retrieved from https://planet.osm.org }.
\newblock \url{ https://www.openstreetmap.org }, 2017.

\bibitem{pang2023detecting}
Chao Pang, Jiang Wu, Jian Ding, Can Song, and Gui-Song Xia.
\newblock Detecting building changes with off-nadir aerial images.
\newblock {\em Science China Information Sciences}, 66(4):140306, 2023.

\bibitem{peng2020semicdnet}
Daifeng Peng, Lorenzo Bruzzone, Yongjun Zhang, Haiyan Guan, Haiyong Ding, and Xu Huang.
\newblock Semicdnet: A semisupervised convolutional neural network for change detection in high resolution remote-sensing images.
\newblock {\em IEEE Transactions on Geoscience and Remote Sensing}, 59(7):5891--5906, 2020.

\bibitem{Priya_fire}
R.S. Priya and K. Vani.
\newblock Vegetation change detection and recovery assessment based on post-fire satellite imagery using deep learning.
\newblock {\em Nature Sci Rep}, 14:12611, 2024.

\bibitem{9420150}
Xiaofan Qu, Feng Gao, Junyu Dong, Qian Du, and Heng-Chao Li.
\newblock Change detection in synthetic aperture radar images using a dual-domain network.
\newblock {\em IEEE Geoscience and Remote Sensing Letters}, 19:1--5, 2022.

\bibitem{ritwik2019xbd}
Gupta Ritwik, Hosfelt Richard, Sajeev Sandra, Patel Nirav, Goodman Bryce, Doshi Jigar, Heim Eric, Choset Howie, and Gaston Matthew.
\newblock xbd: A dataset for assessing building damage from satellite imagery.
\newblock {\em arXiv preprint arXiv:1911.09296}, 2019.

\bibitem{LatentDiffusionModels}
Robin Rombach, A. Blattmann, Dominik Lorenz, Patrick Esser, and Bj{\"o}rn Ommer.
\newblock High-resolution image synthesis with latent diffusion models.
\newblock In {\em CVPR}, 2021.

\bibitem{roscher2023better}
Ribana Roscher, Marc Russwurm, Caroline Gevaert, Michael Kampffmeyer, Jefersson A.~Dos Santos, Maria Vakalopoulou, Ronny Hänsch, Stine Hansen, Keiller Nogueira, Jonathan Prexl, and Devis Tuia.
\newblock Better, not just more: Data-centric machine learning for earth observation.
\newblock {\em IEEE Geoscience and Remote Sensing Magazine}, pages 2--22, 2024.

\bibitem{sachdeva2023change}
Ragav Sachdeva and Andrew Zisserman.
\newblock The change you want to see.
\newblock In {\em WACV}, 2023.

\bibitem{salimans2016improved}
Tim Salimans, Ian Goodfellow, Wojciech Zaremba, Vicki Cheung, Alec Radford, and Xi Chen.
\newblock Improved techniques for training gans.
\newblock {\em Advances in neural information processing systems}, 29, 2016.

\bibitem{SEO_wacv}
Minseok Seo, Hakjin Lee, Yongjin Jeon, and Junghoon Seo.
\newblock Self-pair: Synthesizing changes from single source for object change detection in remote sensing imagery.
\newblock In {\em WACV}, 2023.

\bibitem{SelfPair}
Minseok Seo, Hakjin Lee, Yongjin Jeon, and Junghoon Seo.
\newblock Self-pair: Synthesizing changes from single source for object change detection in remote sensing imagery.
\newblock In {\em WACV}, 2023.

\bibitem{s2looking}
Li Shen, Yao Lu, Hao Chen, Hao Wei, Donghai Xie, Jiabao Yue, Rui Chen, Shouye Lv, and Bitao Jiang.
\newblock S2looking: A satellite side-looking dataset for building change detection.
\newblock {\em Remote Sensing}, 13(24), 2021.

\bibitem{shi2020change}
Wenzhong Shi, Min Zhang, Rui Zhang, Shanxiong Chen, and Zhao Zhan.
\newblock Change detection based on artificial intelligence: State-of-the-art and challenges.
\newblock {\em Remote Sensing}, 12(10):1688, 2020.

\bibitem{singh1989review}
Ashbindu Singh.
\newblock Review article digital change detection techniques using remotely-sensed data.
\newblock {\em International journal of remote sensing}, 10(6):989--1003, 1989.

\bibitem{SyntheWorld}
Jian Song, Hongruixuan Chen, and Naoto Yokoya.
\newblock Syntheworld: A large-scale synthetic dataset for land cover mapping and building change detection.
\newblock In {\em WACV}, 2024.

\bibitem{tang2024changeanywhere}
Kai Tang and Jin Chen.
\newblock Changeanywhere: Sample generation for remote sensing change detection via semantic latent diffusion model.
\newblock {\em arXiv preprint arXiv:2404.08892}, 2024.

\bibitem{Nature_swin}
Yizhuo Tang, Zhengtao Cao, Ningbo Guo, and Mingyong Jiang.
\newblock A {Siamese Swin-Unet} for image change detection.
\newblock {\em Nature Sci Rep}, 14:4577, 2024.

\bibitem{hiucd}
Shiqi Tian, Yanfei Zhong, Ailong Ma, and Zhuo Zheng.
\newblock {Hi-UCD}: A large-scale dataset for urban semantic change detection in remote sensing imagery.
\newblock In {\em NeurIPS Workshop on Machine Learning for the Developing World}, 2020.

\bibitem{TIAN2022164}
Shiqi Tian, Yanfei Zhong, Zhuo Zheng, Ailong Ma, Xicheng Tan, and Liangpei Zhang.
\newblock Large-scale deep learning based binary and semantic change detection in ultra high resolution remote sensing imagery: From benchmark datasets to urban application.
\newblock {\em ISPRS Journal of Photogrammetry and Remote Sensing}, 193:164--186, 2022.

\bibitem{toker2024satsynth}
Aysim Toker, Marvin Eisenberger, Daniel Cremers, and Laura Leal-Taix{\'e}.
\newblock Satsynth: Augmenting image-mask pairs through diffusion models for aerial semantic segmentation.
\newblock In {\em CVPR}, 2024.

\bibitem{Toker_2022_CVPR}
Aysim Toker, Lukas Kondmann, Mark Weber, Marvin Eisenberger, Andr\'es Camero, Jingliang Hu, Ariadna~Pregel Hoderlein, \c{C}a\u{g}lar \c{S}enaras, Timothy Davis, Daniel Cremers, Giovanni Marchisio, Xiao~Xiang Zhu, and Laura Leal-Taix\'e.
\newblock Dynamicearthnet: Daily multi-spectral satellite dataset for semantic change segmentation.
\newblock In {\em CVPR}, 2022.

\bibitem{Van_Etten_2021_CVPR}
Adam Van~Etten, Daniel Hogan, Jesus~Martinez Manso, Jacob Shermeyer, Nicholas Weir, and Ryan Lewis.
\newblock The multi-temporal urban development spacenet dataset.
\newblock In {\em CVPR}, 2021.

\bibitem{WANG2024103761}
Zhipan Wang, Minduan Xu, Zhongwu Wang, Qing Guo, and Qingling Zhang.
\newblock Scribblecdnet: Change detection on high-resolution remote sensing imagery with scribble interaction.
\newblock {\em International Journal of Applied Earth Observation and Geoinformation}, 128:103761, 2024.

\bibitem{xia2023openearthmap}
Junshi Xia, Naoto Yokoya, Bruno Adriano, and Clifford Broni-Bediako.
\newblock {OpenEarthMap}: A benchmark dataset for global high-resolution land cover mapping.
\newblock In {\em WACV}, 2023.

\bibitem{SECOND}
Kunping Yang, Gui-Song Xia, Zicheng Liu, Bo Du, Wen Yang, Marcello Pelillo, and Liangpei Zhang.
\newblock Asymmetric siamese networks for semantic change detection in aerial images.
\newblock {\em IEEE Transactions on Geoscience and Remote Sensing}, 60:1--18, 2022.

\bibitem{zhan2017change}
Yang Zhan, Kun Fu, Menglong Yan, Xian Sun, Hongqi Wang, and Xiaosong Qiu.
\newblock Change detection based on deep siamese convolutional network for optical aerial images.
\newblock {\em IEEE Geoscience and Remote Sensing Letters}, 14(10):1845--1849, 2017.

\bibitem{ControlNet}
Lvmin Zhang, Anyi Rao, and Maneesh Agrawala.
\newblock Adding conditional control to text-to-image diffusion models.
\newblock In {\em ICCV}, 2023.

\bibitem{PerceptualLoss}
Richard Zhang, Phillip Isola, Alexei~A. Efros, Eli Shechtman, and Oliver Wang.
\newblock The unreasonable effectiveness of deep features as a perceptual metric.
\newblock In {\em CVPR}, 2018.

\bibitem{ZHANG20231}
Xiaokang Zhang, Weikang Yu, Man-On Pun, and Wenzhong Shi.
\newblock Cross-domain landslide mapping from large-scale remote sensing images using prototype-guided domain-aware progressive representation learning.
\newblock {\em ISPRS Journal of Photogrammetry and Remote Sensing}, 197:1--17, 2023.

\bibitem{zheng2021nonlocal}
Wen-Jie Zheng, Xi-Le Zhao, Yu-Bang Zheng, and Zhi-Feng Pang.
\newblock Nonlocal patch-based fully connected tensor network decomposition for multispectral image inpainting.
\newblock {\em IEEE Geoscience and Remote Sensing Letters}, 19:1--5, 2021.

\bibitem{zheng2024changen2}
Zhuo Zheng, Stefano Ermon, Dongjun Kim, Liangpei Zhang, and Yanfei Zhong.
\newblock Changen2: Multi-temporal remote sensing generative change foundation model.
\newblock {\em IEEE Transactions on Pattern Analysis and Machine Intelligence}, 2024.

\bibitem{Zheng_2021_ICCV}
Zhuo Zheng, Ailong Ma, Liangpei Zhang, and Yanfei Zhong.
\newblock Change is everywhere: Single-temporal supervised object change detection in remote sensing imagery.
\newblock In {\em ICCV}, 2021.

\bibitem{Changen}
Zhuo Zheng, Shiqi Tian, Ailong Ma, Liangpei Zhang, and Yanfei Zhong.
\newblock Scalable multi-temporal remote sensing change data generation via simulating stochastic change process.
\newblock In {\em ICCV}, 2023.

\bibitem{ZHENG2022228}
Zhuo Zheng, Yanfei Zhong, Shiqi Tian, Ailong Ma, and Liangpei Zhang.
\newblock Changemask: Deep multi-task encoder-transformer-decoder architecture for semantic change detection.
\newblock {\em ISPRS Journal of Photogrammetry and Remote Sensing}, 183:228--239, 2022.

\bibitem{ZHENG2024239}
Zhuo Zheng, Yanfei Zhong, Ji Zhao, Ailong Ma, and Liangpei Zhang.
\newblock Unifying remote sensing change detection via deep probabilistic change models: From principles, models to applications.
\newblock {\em ISPRS Journal of Photogrammetry and Remote Sensing}, 215:239--255, 2024.

\bibitem{ZHU2022113266}
Zhe Zhu, Shi Qiu, and Su Ye.
\newblock Remote sensing of land change: A multifaceted perspective.
\newblock {\em Remote Sensing of Environment}, 282:113266, 2022.

\end{thebibliography}
}

\pagebreak
\clearpage
\setcounter{page}{1}
\setcounter{equation}{0}
\setcounter{section}{0}
\setcounter{figure}{0}
\setcounter{table}{0}

\renewcommand\thefigure{\Alph{figure}}
\renewcommand\thesection{\Alph{section}}
\renewcommand\thetable{\Alph{table}}
\renewcommand\theequation{\Alph{equation}}

\maketitlesupplementary

In this appendix, we provide implementation details in \cref{sec:details}, details about the deep architecture used in \cref{sec:archi_details} and
detailed quantitative results in \cref{sec:quant_results}. Finally, we provide qualitative samples in \cref{sec:qualitatif}.
\begin{figure*}[ht!]
    \centering
     \includegraphics[width=\linewidth]{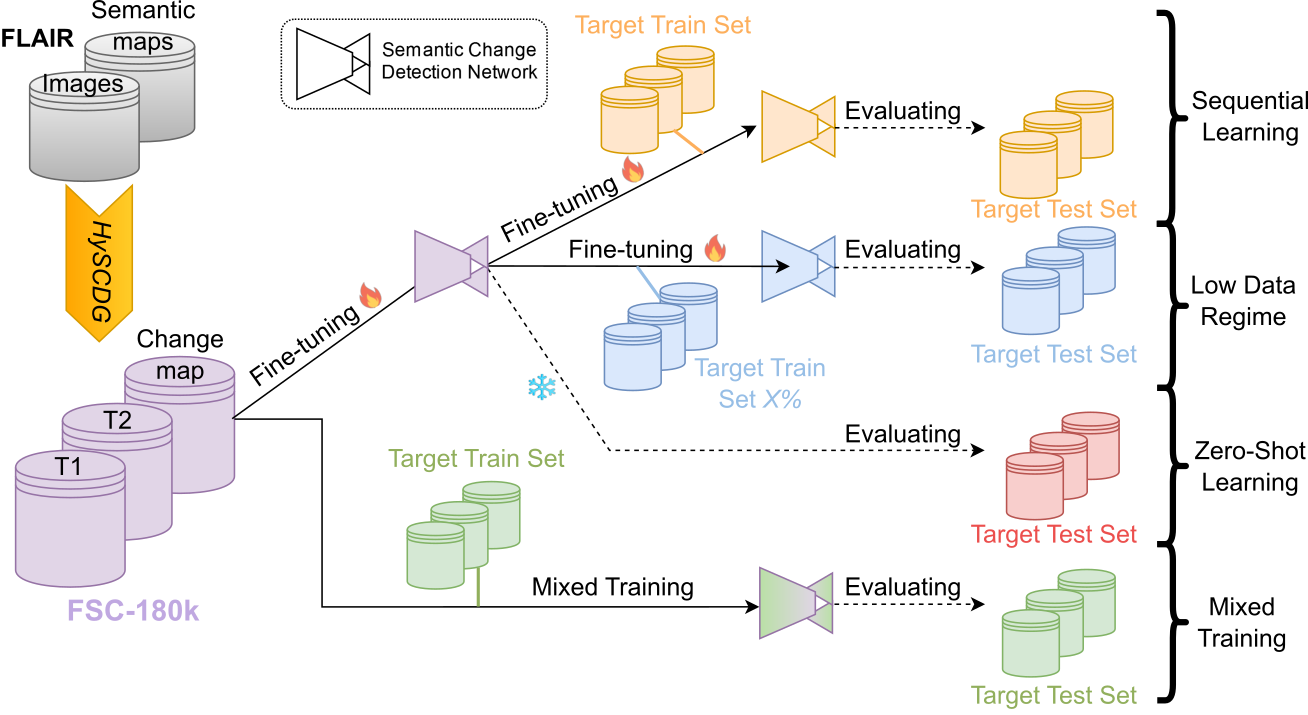}
    \caption{{\bf Transfer Learning scenarios.} We illustrate here the various configurations tested and described in \cref{subsec:tl_expes}.}
    \label{fig:full_pipeline_transfer_Learning}
\end{figure*}

\section{Implementation Details}
\label{sec:details}

\subsection{Training of Stable Diffusion}

For the VAE training, the color loss was applied only after a sufficient amount of 50,000 iterations.
Fine-tuning the VAE took 160 hours A100 GPU for a total of 500,000 iterations with a batch size of 4. 
The diffusion UNet training required 300 hours A100 GPUs, for a total of 30,000 iterations with a batch size of 32. The training was run on 4 GPU in parallel. Finally, we trained the ControlNet for 240 hours on a A100 GPU, performing 45,000 iterations with a batch size of 16. We also observed the \enquote{sudden convergence phenomenon} mentioned in \cite{ControlNet}.

\subsection{The \PIPELINE~pipeline details} %

In the \PIPELINE~pipeline, for a given image with $n$ instances, we randomly select  $N_{change}$  instances (from 0 to $min(3,n)$) to inpaint with a semantic class change and $N_{nochange}$ (from 0 to $min(3,n)$) random shapes for inpainting without semantic class change. We draw the number of instances and shapes from the following heuristical laws (\cref{eq:Nchange,eq:Nnochange}):
\begin{equation}
    N_{change} \sim \left\lfloor \sqrt{\mathcal{U}(0,10)} \right\rfloor,
    \label{eq:Nchange}
\end{equation}
\begin{equation}
   N_{nochange} \sim \left\lfloor \sqrt{\mathcal{U}(0,10) \times (1 - \frac{N_{change}}{4})} \right\rfloor.
   \label{eq:Nnochange}
\end{equation}

The frequency of the inpainted instances and random shapes can be seen in \cref{fig:mask_prob} function to $n$ the number of instances within the image.

\begin{figure*}[h!]
    \centering
    Number of instances in the footprint of a given image
   \includegraphics[width=\linewidth]{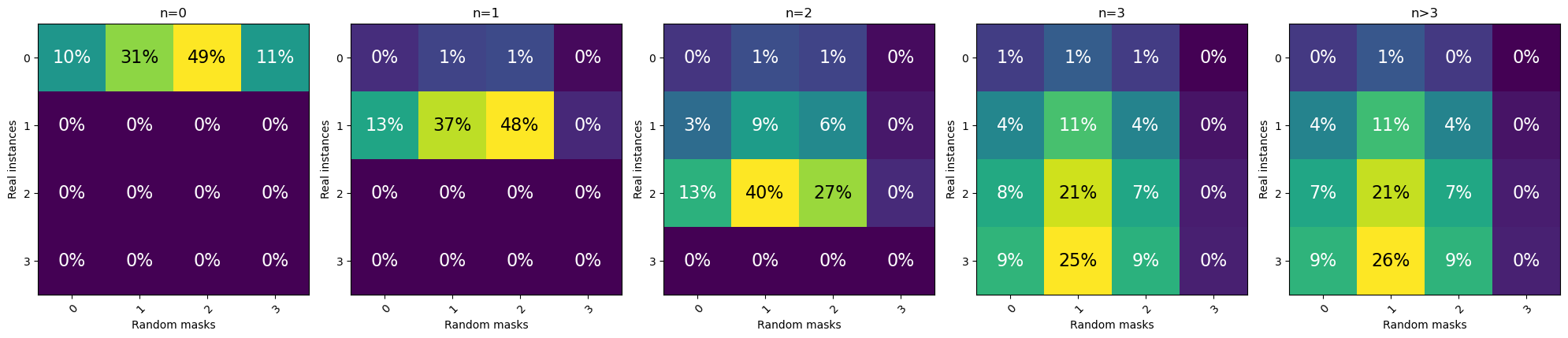}
    \caption{{\bf Heuristic rule to define instance and random mask numbers.} We illustrated the frequency of the number of instances inpainted (rows) and the number of random shapes inpainted (columns) with respect to the number $n$ of instances contained in the footprint of the VHR images.}
    \label{fig:mask_prob}
\end{figure*}

\subsection{FLAIR Dataset \cite{garioud2023flair}}

The FLAIR dataset \cite{garioud2023flair} is composed of 77,762 VHR aerial patches 512 $\times$ 512 at 0.20m GSD. The VHR images include five channels: Red, Green, Blue, near-infrared, and a normalized Digital Surface Model derived by dense image matching (\textit{normalization} means the altitude of the terrain is removed). 
For each patch, ground truth semantic segmentation is provided. The semantic map represents the land cover based on a 19 classes nomenclature (building, coniferous, deciduous, etc. details provided in \cref{fig:generation_pipeline}). We only use the 16 main classes due to the scarcity of certain classes or potential ambiguity. 
This dataset covers approximately 817km$^2$ extracted from various areas in France. 

\subsection{Training computing times}

Models were pretrained to convergence on the synthetic datasets, requiring 40 to 60~hours on a A100 GPU, depending on the configuration. For instance, Dual UNet converges faster than SCanNet, multi-task learning takes longer than binary-only, and the larger size of \FLAIRCHANGE~compared to SyntheWorld requires additional iterations.

For the sequential scenario, fine-tuning was done in 10 to 20~hours (V100 GPU) for HiUCD-mini, Levir-CD, S2Looking and SECOND (in increasing order) and 20~hours (A100 GPU) for HiUCD-XL. For mixed training, all trainings took from 16 to 20~hours (A100 GPU).
Note that, in low data regime, each experiment was repeated 10 times, except for HiUCD-XL for which it was only 3 times for 10\% and 30\% experiments.

\section{Failure cases of the \PIPELINE~synthesis}

ControlNet is the most error-prone module as it may fail to respect the semantic map. Other elements are more reliable or less sensitive: instance footprints of high quality, low influence of text prompt, high-quality inpainting from SD alone  (\Cref{fig:failures_cases}). In this same figure, we show results with only the ControlNet module fine-tuned for generating aerial images from semantic maps, revealing a cartoonish style.

\begin{figure}[ht!]
\vspace{-3mm}
  \centering
\begin{tabular}{cc|cc}
     \multicolumn{2}{c}{Failure cases of \PIPELINE} & \multicolumn{2}{c}{Only ControlNet trained}  \\
     \includegraphics[width=0.1\textwidth]{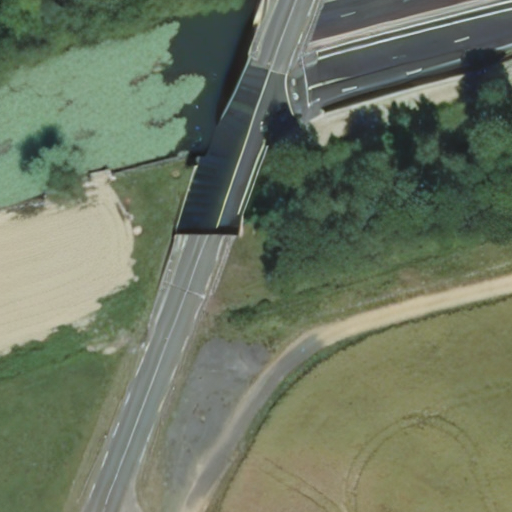} &  \includegraphics[width=0.1\textwidth]{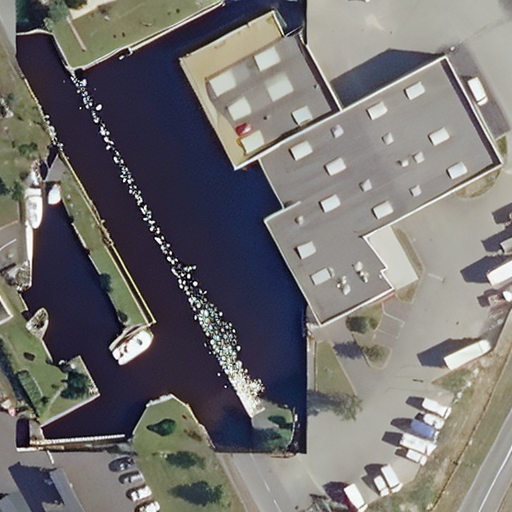} &
     \includegraphics[width=0.1\textwidth]{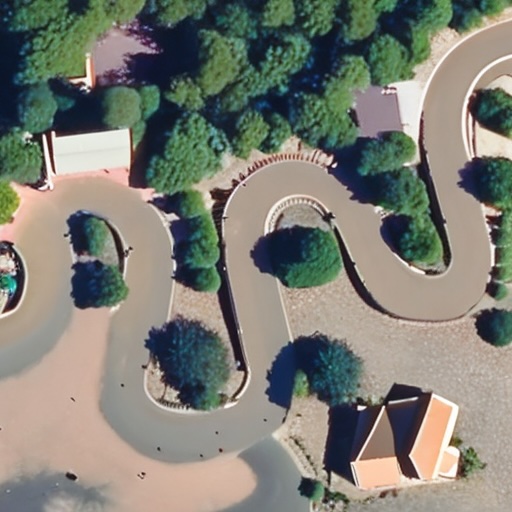} &  \includegraphics[width=0.1\textwidth]{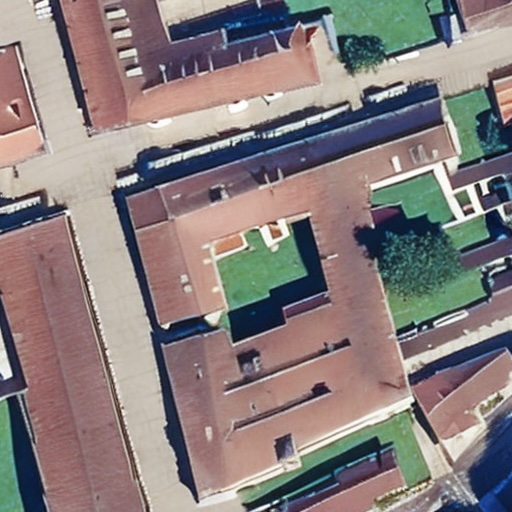}  \\
\end{tabular}
\vspace{-1mm}
\caption{Illustrations of failure cases of synthesis with our pipeline \PIPELINE, and synthesis with only the ControlNet trained.}
\label{fig:failures_cases}
\vspace{-4mm}
\end{figure}

\section{Pretraining and transfer experiments}
\label{sec:archi_details}
\subsection{Transfer Learning Scenarios}

The four transfer learning scenarios considered in this work are illustrated on \cref{fig:full_pipeline_transfer_Learning}. They are quantitatively detailed in Section~\ref{sec:quant_results}.%

\begin{table}[htp!]
    \caption{\textbf{Comparison of model architectures} {\small{The different models were evaluated on SECOND and HiUCD-mini after being trained from scratch for the same duration. \textbf{Bold} values correspond to the best values for each metric and each dataset.}}}
    \label{model_comparison}
    \centering
    \small
    \begin{tabular}{l|l|cc}
    \toprule
         Target & Model & IoU & SeK \\
         \midrule
         \multirow{4}{*}{SECOND} & MambaCD & 0.52 & 0.14  \\
         & SCanNet & \textbf{0.54} & \textbf{0.19} \\
                  & Simple UNet & 0.51 & 0.11  \\
                  & Dual UNet & 0.53 & 0.16 \\
        \midrule 
        \multirow{4}{*}{\begin{tabular}{c}HiUCD  \\ mini\end{tabular}} & MambaCD & 0.53 & 0.08  \\
         & SCanNet & \textbf{0.62} & 0.11 \\
                  & Simple UNet & 0.50 & 0.07  \\
                  & Dual UNet & 0.61 & \textbf{0.17} \\
        \bottomrule  
    \end{tabular}
\end{table}

\subsection{Model architectures}

Our Dual UNet architecture consists of two parallel UNet-style auto-encoders: one dedicated to semantic segmentation for each image separately, and the other one for detecting changes by processing concatenated image pairs. Both UNets share the same core configuration, differing only in their input and output layers: the change detection UNet takes two concatenated images as input and outputs two classes (no-change, change), while the semantic segmentation UNet processes single images and outputs the number of semantic classes (of the dataset).
Both networks use a ResNet-50 encoder pretrained on the ImageNet dataset. To enhance performance, features extracted at each encoding step of the semantic UNet are injected into the decoding pathway of the change detection UNet, in addition to its own skip connection features.
The entire model comprises a total of 83 million parameters.

We compared our architecture with state-of-the-art change detection models that rely on different mechanisms: MambaCD \cite{ChangeMamba}, SCanNet \cite{SCanNet}, and ChangeFormer \cite{ChangeFormer}. The latter, ChangeFormer was only evaluated through pretraining on \FLAIRCHANGE, which proved unsuccessful due to the model's excessive size and the significant computational resources required for training. MambaCD and SCanNet were evaluated on SECOND and HiUCD-mini. The performances of the different models can be found in \cref{model_comparison}. SCanNet was better in all cases and was therefore kept for comparison in sequential experiments, with \FLAIRCHANGE~pretraining.

\subsection{Normalization}
We compared the effects of applying or not normalization by mean and variance of the train set pixel's values on the input data. It proved to be really effective. For example, this improves SCanNet's performance on the SECOND dataset from 0.17 to 0.19 in Separated Kappa. We used it in all our experiments, using the parameters of the handled dataset. Namely, in sequential mode, normalization uses the synthetic dataset's parameters during pretraining, and the target dataset parameters during fine-tuning. 

\begin{table}[htp]
    \caption{\textbf{Effect of the normalization.} {\small{The impact of input normalization was evaluated on the SECOND dataset. For both SCanNet and Dual UNet, normalization was highly effective.}}}
    \label{normalization_second}
    \centering
    \small
    \begin{tabular}{l|c|cc}
    \toprule
         Model & Normalization & IoU & SeK \\
         \midrule
         \multirow{2}{*}{SCanNet} &  & 0.52 & 0.17  \\
         & X & \textbf{0.54} & \textbf{0.19} \\
         \midrule
          \multirow{2}{*}{Dual UNet}         &  & 0.51 & 0.14  \\
                  & X & \textbf{0.53} & \textbf{0.16} \\
        \bottomrule  
    \end{tabular}
\end{table}

\section{Quantitative results}
\label{sec:quant_results}

In this section, we present in details the quantitative results for both binary and semantic cases, for the four transfer learning scenarios. We follow the same order as in the paper.

\begin{compactitem}
\item \textbf{Sequential Training}: binary change detection (BCD) results are provided in \cref{tab:sequential_bcd}, while semantic change detection (SCD) outputs are given in \cref{tab:sequential_scd}.
\item \textbf{Low-Data Regime}: results for BCD can be found in \cref{tab:low_data_regime_bcd} for Levir-CD et S2Looking, and in \cref{tab:low_data_regime_scd} for SECOND and HiUCD-mini for the SCD configuration.
\item \textbf{Mixed Training}: \cref{tab:mixed_bcd} and \cref{tab:mixed_scd} provide numbers, respectively, for BCD for the four datasets and SCD for SECOND and HiUCD-mini datasets. \cref{tab:mixed_scd_hiucdfull} focuses on HiUCD-XL dataset.
\item \textbf{Zero-Shot Learning}: results for BCD and SCD are given in \cref{tab:zeroshot_bcd} and \cref{tab:zeroshot_scd} respectively.
\end{compactitem}

\vspace{0.5cm}
\subsection{Binary change detection}

\begin{table}[h!]
        \caption{\textbf{Sequential Training Evaluation (Binary Change Detection).} We report the F1 and IoU scores on four datasets. Models were initially pretrained ({\color{synthecolor}{SyntheWorld}} or {\color{flaircolor}{\FLAIRCHANGE}}) or not ({\color{baselinecolor}{Baseline}}) and then fine-tuned on the target dataset. Pretraining effects remain limited on simpler datasets such as Levir-CD, but provide significant benefits for more complex datasets like HiUCD, which is also closer to our \FLAIRCHANGE~in terms of landscapes and characteristics. \textbf{Bold} values correspond to the best values for each metric and each dataset.}
    \label{tab:sequential_bcd}
    \centering
    \setlength{\tabcolsep}{4pt}
    \small
    \begin{tabular}{ll|cccc}
    \toprule
         & \begin{tabular}{c}
            Target $	\rightarrow$ \\
              Pretraining $\downarrow$
         \end{tabular}    & \rotatebox{70}{Levir-CD} & \rotatebox{70}{S2Looking} & \rotatebox{70}{SECOND} & \rotatebox{70}{\begin{tabular}{c}HiUCD  \\ mini\end{tabular}} \\
         \midrule
       \multirow{3}{*}{F1} & {\color{baselinecolor}Baseline} & \textbf{0.91} & 0.61 & 0.70 & 0.75 \\
           & {\color{synthecolor}SyntheWorld} & \textbf{0.91} & \textbf{0.63} & \textbf{0.71} & 0.73  \\
           & {\color{flaircolor}\FLAIRCHANGE} & \textbf{0.91} & \textbf{0.63} & \textbf{0.71} & \textbf{0.77} \\
      \midrule
           \multirow{3}{*}{IoU} & {\color{baselinecolor}Baseline} & 0.83 & 0.44 & 0.53 & 0.61 \\
           & {\color{synthecolor}SyntheWorld} & \textbf{0.84} & \textbf{0.46} & 0.54 & 0.58 \\
           & {\color{flaircolor}\FLAIRCHANGE} & \textbf{0.84} & \textbf{0.46} & \textbf{0.55} & \textbf{0.63}  \\
    \bottomrule
    \end{tabular}

\end{table}

\FloatBarrier

\begin{table}[h!]
 \caption{\textbf{Low Data Regime (Binary Change Detection).} We report the IoU scores on two binary datasets. Models were initially pretrained ({\color{synthecolor}{SyntheWorld}} or {\color{flaircolor}{\FLAIRCHANGE}}) or not ({\color{baselinecolor}{Baseline}}) and then fine-tuned on a portion of the target dataset (X\%). Pretraining offers significant advantages in scenarios with limited data, making it particularly valuable in many real-world applications. \textbf{Bold} values correspond to the best values for each metric and each dataset.}
     \label{tab:low_data_regime_bcd}
    \centering
    \setlength{\tabcolsep}{4pt}
    \small
    \begin{tabular}{l|lll}
    \toprule
         Target Percent & 1 \% & 10 \% & 30 \% \\
           \midrule
          \multicolumn{4}{c}{Levir-CD} \\
         \midrule
         {\color{baselinecolor}Baseline} & 0.36  & 0.69 & 0.75     \\
         {\color{synthecolor}SyntheWorld} & 0.41 (+14\%)  & 0.71 (+3\%) & 0.77 (+3\%)    \\
         {\color{flaircolor}\FLAIRCHANGE} & \textbf{0.55} (+53\%)  & \textbf{0.73} (+6\%) & \textbf{0.79} (+5\%)    \\
         \midrule
         \multicolumn{4}{c}{S2Looking} \\
         \midrule
          {\color{baselinecolor}Baseline} & 0.10   & 0.27  & 0.38   \\ 
          {\color{synthecolor}SyntheWorld} & 0.08 (-20\%)  & 0.34 (+26\%)  & 0.41 (+8\%)    \\
         {\color{flaircolor}\FLAIRCHANGE} & \textbf{0.15} (+50\%) & \textbf{0.36} (+33\%) & \textbf{0.43} (+13\%)    \\
         \bottomrule
    \end{tabular}
\end{table}

\FloatBarrier

\begin{table}[h!]
    \caption{\textbf{Mixed Training Evaluation (Binary Change Detection).} Model are trained on a blend of target and synthetic/hybrid ({\color{synthecolor}SyntheWorld} or {\color{flaircolor}\FLAIRCHANGE}) train sets, containing a ratio of x\% samples from the target (including repetitions). Testing is performed on the target test set. The last column corresponds to train exclusively on target dataset (without pretraining).  We report the F1 and IoU scores on four datasets. Mixed training turns out to be at least as effective as sequential one. \textbf{Bold} values correspond to the best values for each metric and each dataset.}
    \label{tab:mixed_bcd}
    \centering
          \setlength{\tabcolsep}{4pt}
          \small
    \begin{tabular}{l|c|l|cccc}
       \toprule
       Target   & & Pretraining  & 20\% & 50\% & 90\% & 100\% \\
         \midrule
         \multirow{4}{*}{Levir-CD}  & \multirow{2}{*}{F1} & {\color{synthecolor}SyntheWorld} &0.90 & 0.90 & \textbf{0.91} & \multirow{2}{*}{0.90} \\
         & &  {\color{flaircolor}\FLAIRCHANGE} & 0.90 & \textbf{0.91} & \textbf{0.91} & \\
                   & \multirow{2}{*}{IoU} & {\color{synthecolor}SyntheWorld} & 0.81 & 0.83 & \textbf{0.84} & \multirow{2}{*}{0.83} \\
                   & &  {\color{flaircolor}\FLAIRCHANGE} & 0.82 & 0.83 & \textbf{0.84} &   \\
        \midrule
         \multirow{4}{*}{S2Looking} & \multirow{2}{*}{F1}& {\color{synthecolor}SyntheWorld} & 0.60 & 0.63 & 0.62 & \multirow{2}{*}{0.61} \\
          & &  {\color{flaircolor}\FLAIRCHANGE} & 0.61 & \textbf{0.64} & 0.63 & \\
                   & \multirow{2}{*}{IoU} & {\color{synthecolor}SyntheWorld} & 0.43 & 0.48 & 0.48 & \multirow{2}{*}{0.44} \\
                   & &  {\color{flaircolor}\FLAIRCHANGE}   & 0.44 & \textbf{0.49} & \textbf{0.49} & \\
        \hline
         \multirow{4}{*}{SECOND}  & \multirow{2}{*}{F1} & {\color{synthecolor}SyntheWorld} & 0.71 & 0.72 & 0.71 & \multirow{2}{*}{0.70} \\
                                &  &  {\color{flaircolor}\FLAIRCHANGE}& 0.72 & \textbf{0.73} & 0.72 &  \\
                   & \multirow{2}{*}{IoU} & {\color{synthecolor}SyntheWorld} & 0.54 & 0.56 & 0.55 & \multirow{2}{*}{0.53}  \\
                   & &  {\color{flaircolor}\FLAIRCHANGE}  & \textbf{0.57} & \textbf{0.57} & 0.55 &   \\
        \hline
         \multirow{4}{*}{\begin{tabular}{c}HiUCD  \\ mini\end{tabular}}  & \multirow{2}{*}{F1} & {\color{synthecolor}SyntheWorld}&0.71 & 0.72 & 0.72 & \multirow{2}{*}{0.75} \\
                                & &  {\color{flaircolor}\FLAIRCHANGE} & 0.76 & 0.76 & \textbf{0.77} &   \\
                   & \multirow{2}{*}{IoU} & {\color{synthecolor}SyntheWorld} & 0.58 & 0.58 & 0.59 & \multirow{2}{*}{0.61} \\
                   &  &  {\color{flaircolor}\FLAIRCHANGE}& 0.61 & 0.62 & \textbf{0.63} &  \\
           \bottomrule
    \end{tabular}
\end{table}

\begin{table}[!ht]
    \caption{\textbf{Zero-Shot Evaluation (Binary Change Detection).} We report the F1 and IoU scores for the zero-shot case. Metrics for SECOND and HiUCD are available on \cref{tab:zeroshot_scd}. Models were initially pretrained ({\color{synthecolor}{SyntheWorld}} or {\color{flaircolor}{\FLAIRCHANGE}}). Decent performance was obtained on Levir-CD, despite the domain gap between \FLAIRCHANGE~and Levir-CD. However, pretraining remained completely ineffective for S2Looking. \textbf{Bold} values correspond to the best values for each metric and each dataset.}
    \label{tab:zeroshot_bcd}
    \small
    \centering
    \begin{tabular}{l|l|ccc}
        \toprule
         Target & Pretraining & F1 & IoU \\
         \midrule
         \multirow{2}{*}{Levir-CD} %
         & {\color{synthecolor}{SyntheWorld}} & 0.25 & 0.13  \\
                 & {\color{flaircolor}{\FLAIRCHANGE}} & \textbf{0.49} & \textbf{0.33 } \\
        \midrule
        \multirow{2}{*}{S2Looking} %
         & {\color{synthecolor}{SyntheWorld}} & 0.0 & 0.0  \\
                 & {\color{flaircolor}{\FLAIRCHANGE}} & 0.04 & 0.02  \\
       \bottomrule
    \end{tabular}
\end{table}

\newpage
~~~~~
\newpage
\subsection{Semantic change detection}
\begin{table}[ht!]
    \caption{\textbf{Sequential Training Evaluation (Semantic Change Detection).} We report the binary and semantic scores on three datasets. Models (based on the Dual UNet) were initially pretrained ({\color{synthecolor}{SyntheWorld}} or {\color{flaircolor}{\FLAIRCHANGE}}) or not ({\color{baselinecolor}{Baseline}}), and, then, fine-tuned on the target dataset. Pretraining on \FLAIRCHANGE~proved highly effective and outperformed SyntheWorld on every semantic dataset, with greater benefits observed on HiUCD due to the higher similarity between the datasets. \textbf{Bold} values correspond to the best values for each metric and each dataset.}
    \label{tab:sequential_scd}
    \centering
    \setlength{\tabcolsep}{3pt}
    \small
    \begin{tabular}{l|l|ccccc}
        \toprule
         Target & Pretraining  & IoU  & Ovr. IoU & SeK & SCS \\
         \midrule
                 \multirow{3}{*}{\rotatebox{50}{SECOND}} &  {\color{baselinecolor}Baseline}  & 0.53  & 0.64 & 0.16 & 0.83 \\
         & {\color{synthecolor}SyntheWorld} & 0.54  & 0.63 & 0.17 & 0.84 \\
        &  {\color{flaircolor}{\FLAIRCHANGE}} & \textbf{0.55} & \textbf{0.65} & \textbf{0.18} & \textbf{0.89} \\
        \midrule
        \multirow{3}{*}{
            \rotatebox{50}{\begin{tabular}{c}HiUCD  \\ mini\end{tabular}}} & {\color{baselinecolor}Baseline}  & 0.61  & 0.77 & 0.17 & 0.73 \\
         &{\color{synthecolor}SyntheWorld} & 0.58  & 0.78 & 0.15 & 0.73 \\
         &{\color{flaircolor}\FLAIRCHANGE} & \textbf{0.63}  & \textbf{0.79} & \textbf{0.19} & \textbf{0.78} \\
         \midrule \midrule
       & & \begin{tabular}{c}Sem.  \\ mIoU\end{tabular} &  \begin{tabular}{c}Chg.  \\ mIoU\end{tabular}  &  \begin{tabular}{c}Bin.  \\ mIoU\end{tabular}  &  \begin{tabular}{c}Bin. C. \\ mIoU\end{tabular} \\
        \midrule
          \multirow{3}{*}{\rotatebox{50}{\begin{tabular}{c}HiUCD  \\ XL\end{tabular}}} & {\color{baselinecolor}Baseline}   & 0.58 & 0.17  & 0.34 & \textbf{0.48} \\
        & {\color{synthecolor}SyntheWorld}  & 0.58 & 0.17  & 0.34 & 0.48 \\
        & {\color{flaircolor}{\FLAIRCHANGE}} & \textbf{0.60} & \textbf{0.19}  & 0.34 & \textbf{0.48} \\
    \bottomrule
    \end{tabular}
\end{table}

\FloatBarrier

\begin{table}[htp]
     \caption{\textbf{Low Data Regime (Semantic Change Detection).} Models were initially pretrained ({\color{synthecolor}{SyntheWorld}} or {\color{flaircolor}{\FLAIRCHANGE}}) or not ({\color{baselinecolor}{Baseline}}) and then fine-tuned on a portion of the target dataset (x\%). The benefit of pretraining increases as the amount of available data decreases. These experiments highlight the relevance of pretraining on our \FLAIRCHANGE~in real-world cases where annotated data is often scarce and/or costly to obtain. \textbf{Bold} values correspond to the best values for each metric and each dataset.}
     \label{tab:low_data_regime_scd}
    \centering
    \setlength{\tabcolsep}{4pt}
    \small
    \begin{tabular}{l|lll}
    \toprule
          Target Percent & 1 \% & 10 \% & 30 \% \\
           \midrule
          \multicolumn{4}{c}{SECOND (scores in SCS)} \\
         \midrule
         {\color{baselinecolor}Baseline} & 0.31  & 0.49 & 0.69     \\
         {\color{synthecolor}SyntheWorld} & 0.37 (+18\%)  & 0.49 (+0\%)  & 0.64 (-6\%)    \\
         {\color{flaircolor}\FLAIRCHANGE} & \textbf{0.40} (+29\%)  & \textbf{0.57} (+15\%) & \textbf{0.74} (+8\%)    \\
         \midrule
         \multicolumn{4}{c}{HiUCD-mini (scores in SCS)} \\
         \midrule
          {\color{baselinecolor}Baseline} & 0.13   & 0.41  & 0.48   \\ 
          {\color{synthecolor}SyntheWorld} & 0.17 (+31\%)  & 0.38 (-7\%)  & 0.50 (+4\%)    \\
         {\color{flaircolor}\FLAIRCHANGE} & \textbf{0.20} (+38\%) & \textbf{0.49} (+19\%) & \textbf{0.54} (+13\%)    \\
         \midrule
         \multicolumn{4}{c}{HiUCD-XL (scores in Chg. mIoU)} \\
         \midrule
          {\color{baselinecolor}Baseline} & 0.06   & 0.10  & 0.12   \\ 
          {\color{synthecolor}SyntheWorld} & 0.05 (-16\%)  & 0.10 (+0\%) & 0.11 (-8\%)    \\
         {\color{flaircolor}\FLAIRCHANGE} & \textbf{0.08} (+33\%) & \textbf{0.13} (+30\%) & \textbf{0.15} (+25\%)    \\
         \bottomrule
    \end{tabular}
\end{table}

\FloatBarrier
\begin{table}[!ht]
    \caption{\textbf{Mixed Training Evaluation (Semantic Change Detection).} Model are trained on a blend of target and synthetic/hybrid ({\color{synthecolor}SyntheWorld} or {\color{flaircolor}\FLAIRCHANGE}) train sets, containing a ratio of x\% samples from the target (including repetitions). Testing is performed on the target test set. The last column corresponds to train exclusively on target dataset (without pretraining).  Mixed training turns out to be more effective than sequential training, thanks to the exposure of the model to diverse and varied examples throughout the training process. \textbf{Bold} values correspond to the best values for each metric and each dataset.}
    \label{tab:mixed_scd}
    \centering
    \setlength{\tabcolsep}{4pt}
      \small
    \begin{tabular}{l|c|l|cccc}
    \toprule
         Target   & & Pretraining & 20\% & 50\% & 90\% & 100\% \\
         \midrule
         \multirow{8}{*}{\rotatebox{90}{SECOND}}  & \multirow{2}{*}{IoU} & {\color{synthecolor}SyntheWorld} & 0.54 & 0.56 & 0.55 & \multirow{2}{*}{0.53} \\
         & & {\color{flaircolor}\FLAIRCHANGE} & 0.57 & \textbf{0.57}& 0.55 &  \\
                   & \multirow{2}{*}{Ovr. IoU}  & {\color{synthecolor}SyntheWorld} & 0.61 & 0.62 & 0.63 & \multirow{2}{*}{0.64} \\
                   & & {\color{flaircolor}\FLAIRCHANGE} & \textbf{0.64} & 0.63 & 0.62 &  \\
                   & \multirow{2}{*}{SeK}& {\color{synthecolor}SyntheWorld}  & 0.16 & 0.17 & 0.16 & \multirow{2}{*}{0.16} \\
                   & & {\color{flaircolor}\FLAIRCHANGE} & \textbf{0.19} & 0.18 & 0.18 &  \\
                   & \multirow{2}{*}{SCS} & {\color{synthecolor}SyntheWorld} & 0.83 & 0.84 & 0.85 & \multirow{2}{*}{0.83} \\
                   & & {\color{flaircolor}\FLAIRCHANGE} & 0.88 & \textbf{0.89} & 0.88 &  \\
        \midrule
         \multirow{8}{*}{\rotatebox{90}{\begin{tabular}{c}HiUCD  \\ mini\end{tabular}}}  & \multirow{2}{*}{IoU}& {\color{synthecolor}SyntheWorld}  & 0.58 & 0.58 & 0.59 & \multirow{2}{*}{0.61}\\
                                & & {\color{flaircolor}\FLAIRCHANGE} & 0.61 & 0.62 & \textbf{0.63} & \\
                   & \multirow{2}{*}{Ovr. IoU}& {\color{synthecolor}SyntheWorld}  & 0.74 & \textbf{0.78} & 0.77 & \multirow{2}{*}{0.77} \\
                   & & {\color{flaircolor}\FLAIRCHANGE} & 0.76 & 0.77 & 0.77 &  \\
                   & \multirow{2}{*}{SeK}& {\color{synthecolor}SyntheWorld}  & 0.11 & 0.13 & 0.14 & \multirow{2}{*}{0.17}\\
                   & & {\color{flaircolor}\FLAIRCHANGE} & 0.16 & 0.17 & \textbf{0.18} & \\
                   & \multirow{2}{*}{SCS}& {\color{synthecolor}SyntheWorld}  & 0.68 & 0.71 & 0.72 & \multirow{2}{*}{0.73} \\
                   & & {\color{flaircolor}\FLAIRCHANGE} & 0.74 & 0.76 & \textbf{0.78} &  \\
        \bottomrule
    \end{tabular}
\end{table}

\begin{table}[!ht]
    \caption{\textbf{Mixed Training Evaluation on HiUCD-XL.}  Training on a blend of target and synthetic/hybrid ({\color{synthecolor}SyntheWorld} or {\color{flaircolor}\FLAIRCHANGE}) train sets, containing a ratio of x\% samples from the target (including repetitions). Testing is performed on the target test set. 100\% corresponds to fine-tuning exclusively on target dataset (without pretraining). 0\% corresponds to zero-shot (after pretraining). An optimal mix ratio appears to be around 50\%, which can be understood as a good compromise between exploration (pretraining data) and exploitation (target data). \textbf{Bold} values correspond to the best values for each metric.}
    \label{tab:mixed_scd_hiucdfull}
    \centering
     \setlength{\tabcolsep}{4pt}
    \small
    \begin{tabular}{ll|ccccc}
    \toprule
         Pretraining & Mix ratio & \begin{tabular}{c}Sem.  \\ mIoU\end{tabular} & \begin{tabular}{c}Chg.  \\ mIoU\end{tabular} & \begin{tabular}{c}Bin.  \\ mIoU\end{tabular}& \begin{tabular}{c}Bin. C.  \\ mIoU\end{tabular} \\
         \midrule
        \multirow{4}{*}{\rotatebox{50}{\color{synthecolor}{SyntheWorld}}} & 0\%  & 0.02 & 0.03  & 0.0 & 0.47 \\
         & 20\%  & 0.51 & 0.10  & 0.33 & 0.48 \\
        & 50\% & 0.62 & 0.19  &\textbf{ 0.37} & 0.51 \\
         & 90\% & 0.60 & 0.17  & 0.32 & 0.47 \\
        \midrule
        \multirow{4}{*}{\rotatebox{50}{\color{flaircolor}{\FLAIRCHANGE}}} & 0\%  & 0.35 & 0.07  & 0.20 & 0.43 \\
         & 20\%  & 0.58 & 0.15  & 0.32 & 0.51 \\
         & 50\% & \textbf{0.63} & \textbf{0.20}  & 0.29 & \textbf{0.52} \\
        & 90\% & 0.62 & 0.18  & 0.30 & 0.51 \\
        \midrule
      - & 100\% & 0.58 & 0.17  & 0.34 & 0.48 \\
        \bottomrule
    \end{tabular}
\end{table}

\begin{table}[!h]
 \caption{\textbf{Zero-Shot Evaluation (Semantic Change Detection)}. Models were initially pretrained ({\color{synthecolor}{SyntheWorld}} or {\color{flaircolor}{\FLAIRCHANGE}}). In semantic mode, only \FLAIRCHANGE~pretraining enables the model to detect changes and predict semantic segmentation. \textbf{Bold} values correspond to the best values for each metric and each dataset.}
    \label{tab:zeroshot_scd}
    \centering
     \setlength{\tabcolsep}{4pt}
    \small
    \begin{tabular}{l|l|ccc}
    \toprule
         Target & Pretraining & F1 & IoU & SCS \\
         \midrule
         \multirow{2}{*}{SECOND} 
         & {\color{synthecolor}{SyntheWorld}} & 0.02 & 0.01 & 0.0 \\
                  & {\color{flaircolor}{\FLAIRCHANGE}} & \textbf{0.53} & \textbf{0.36} &  \textbf{0.24} \\
        \midrule
         \multirow{2}{*}{\begin{tabular}{c}HiUCD  \\ mini\end{tabular}} 
         & {\color{synthecolor}{SyntheWorld}} & 0.02 & 0.01 & 0.0 \\
                 & {\color{flaircolor}{\FLAIRCHANGE}} & \textbf{0.53} & \textbf{0.36} & \textbf{0.25}  \\
        \midrule
         &  & Sem. mIoU & Chg mIoU & Bin. mIoU \\
         \midrule
         \multirow{2}{*}{\begin{tabular}{c}HiUCD  \\ XL\end{tabular}} & {\color{synthecolor}{SyntheWorld}} & 0.02 & 0.03 & 0.0 \\
                 & {\color{flaircolor}{\FLAIRCHANGE}} & \textbf{0.35} & \textbf{0.07} & \textbf{0.20}  \\
        \bottomrule   
    \end{tabular}
\end{table}

\clearpage
\onecolumn
\section{Qualitative results}
\label{sec:qualitatif}
In \cref{fig:FLAIRChange_samples}, we provide some examples of images generates with \PIPELINE, that we extracted from our dataset  \FLAIRCHANGE. One can see the joint spatial and semantic accuracy of the images and maps, coupled with diverse and real-world change configurations. The random selection of instances allows to provide multiple change trajectories and not to focus on main classes (namely, \textit{Building}, \textit{Impervious surfaces}, and \textit{Bare soil}). The controlled inpainting is fairly realistic, for example with the addition of an agricultural field to replace a sport ground at the top right of the first line image.
\input{tikz_fig/Generated_samples}
\clearpage
\FloatBarrier
\cref{fig:second_samples} and \cref{fig:hiucd_samples} provide several examples of semantic prediction results with our Dual UNet architecture for SECOND and HiUCD-mini datasets, respectively. Experiments were performed in sequential scenario: without synthetic nor hybrid training data (Baseline), pretraining on SyntheWorld, and pretraining on \FLAIRCHANGE.\\
One can first see that the Dual UNet architecture alone is sufficient for reliable dense predictions on VHR images, validating our quantitative assessment related to model architectures (\cref{model_comparison}). 
Secondly, it can be noticed that working with \FLAIRCHANGE~allows for more consistent results over various classes, landscapes, and datasets. It is also worth noting that the GSD varies between the pretraining dataset (0.2m) and the target datasets (0.1m for HiUCD and approximately 0.45m for SECOND).
\input{tikz_fig/Second_change2}

\input{tikz_fig/HiUCD_change}

\end{document}